\documentclass[12pt,a4paper]{article}

\usepackage[margin=0.8in]{geometry}

\usepackage[utf8]{inputenc}
\usepackage{etoolbox}
\usepackage{amsfonts}
\usepackage{amssymb}
\usepackage[square,numbers]{natbib}
\usepackage{hyperref}
\usepackage{listings}
\usepackage{pdfpages}
\usepackage{float}
\usepackage{makecell}
\usepackage{url}
\usepackage{pgfplots}
\pgfplotsset{compat=newest}
\usepackage{subfigure}

\usepackage[numbib]{tocbibind}

\usepackage{graphicx}
\usepackage{color}
\usepackage{amsmath}
\usepackage{times}
\usepackage{soul}
\usepackage{url}
\usepackage[utf8]{inputenc}
\usepackage[small]{caption}
\usepackage{booktabs}
\usepackage{algorithmic}
\usepackage[linesnumbered,ruled]{algorithm2e}
\usepackage{mathtools, cuted}
\usepackage{multicol, multirow}
\usepackage[]{xcolor}
\usepackage{appendix}
\usepackage{caption}
\usepackage{setspace}

\graphicspath{ {images/} }

\begin{document}

\begin{titlepage}
	\begin{center}
		\vspace{0.5cm}
		\vspace{0.5cm}
		\hrule
		\vspace{3cm}
		\normalsize{Manuscript}\\
		\vspace{0.5cm}
		\LARGE{A prediction-based approach \\ for online dynamic
			patient scheduling: \\
			a case study in radiotherapy treatment}\\
		\vspace{3cm}
		\Large{Tu-San Pham$^{a,c,*}$, Antoine Legrain$^{a,c}$,} \\
		\Large{Patrick De Causmaecker$^b$, Louis-Martin Rousseau$^{a,c}$ }\\
		\vspace{0.5cm}
		\normalsize{\today}\\
		\vspace{3cm}
		\begin{tabular}{l}
			$^*$ Corresponding author. \\ 
			Email: 
			tu-san.pham@polymtl.ca  \\ \\
			Affiliations:\\
			$^a$ Polytechnique Montréal, Quebec, Canada  \\  %
			$^b$ KU Leuven, Kortrijk, Belgium \\
			$^c$ CIRRELT
		\end{tabular}		
	\end{center}
\end{titlepage}

\section{Abstract}

	Patient scheduling is a difficult task involving stochastic factors such as the unknown arrival times of patients. Similarly, the scheduling of radiotherapy for cancer treatments needs to handle patients with different urgency levels when allocating resources. High priority patients may arrive at any time, and there must be resources available to accommodate them. A common solution is to reserve a flat percentage of treatment capacity for emergency patients. However, this solution can result in overdue treatments for urgent patients, a failure to fully exploit treatment capacity, and delayed treatments for low-priority patients. This problem is especially severe in large and crowded hospitals. In this paper, we propose a prediction-based approach for online dynamic radiotherapy scheduling that dynamically adapts the present scheduling decision based on each incoming patient and the current allocation of resources. Our approach is based on a regression model trained to recognize the links between patients' arrival patterns, and their ideal waiting time in optimal offline solutions where all future arrivals are known in advance. 
	When our prediction-based approach is compared to flat-reservation policies, it does a better job of preventing overdue treatments for emergency patients, while also maintaining comparable waiting times for the other patients. We also demonstrate how our proposed approach supports explainability and interpretability in scheduling decisions using SHAP values. \\

Keywords: Operations Research; Radiotherapy scheduling; Integer Programming; Patient Scheduling; Explainability;


\section{Introduction}
According to the WHO \footnote{\url{https://www.who.int/news-room/fact-sheets/detail/cancer}}, cancer is the leading cause of death worldwide, responsible for nearly 10 million deaths in 2020. In Canada, the number of cancer deaths in 2020 was 83,300, accounting for about $30\%$ of all deaths in the country. 
The number of cancer incidents is steadily rising in many countries, which leads to increased stress on treatment facilities and medical staff, while also making it more difficult to get timely access to necessary cancer treatments.  
Numerous studies \citep{coles2003audit,Mackillop07,chen2008relationship} have proven that a long waiting time preceding cancer treatment has a negative impact on the clinical outcome of that treatment (e.g. persistence of cancer symptoms, psychological distress, increase of death rate). In cancer treatment, waiting time is referred to as killing time \citep{Mackillop07}. 
Therefore, it is essential to improve the treatment scheduling process, as it will reduce the waiting time before patients start treatment.\looseness=-1  

In this paper, we focus on radiotherapy (RT), one of the most popular forms of cancer treatment. About $50\%$ of cancer patients undergo RT treatment \citep{barton2014estimating,tyldesley2011estimating}. During RT treatment, a patient receives a high dose of radiation divided into small portions (called \textit{fractions}) that are delivered to the cancer site over several consecutive days. 
This treatment is most commonly delivered by a linear accelerator (\textit{linac}). 
As the number of linacs in a hospital is usually limited, the waiting time for cancer patients to start their RT treatments directly depends on treatment scheduling.  
We consider the RT scheduling problem arising at CHUM (Centre hospitalier de l'Universit\'e de Montr\'eal), a large cancer center in Montr\'eal, Canada. At CHUM, patients are categorized by the maximum recommended waiting time given by their doctor. Overdue treatments (treatments started after their recommended deadline) are highly discouraged. 
The objective is to minimize overdue treatments and patient waiting times. 
The main challenge presented by this problem is uncertainty in the demand for treatment by patients with different urgency levels. 
		Palliative patients need urgent care to relieve intense pain, and so their treatment deadline is set to 1 to 3 days after their admission. In contrast, curative patients can wait for 2 to 4 weeks. 
Approximately $70\%$ of patients treated at CHUM are curative. Despite accounting for only $30\%$ of treatments, palliative treatments are the most challenging group to schedule due to their late arrivals and short treatment deadlines \citep{pham2021two}. 
	Scheduling curative patients too early risks leading to overdue treatments in palliative patients by leaving too few reserved treatment slots available. However, delaying treatments in curative patients for too long could lead to inefficient capacity usage, while also increasing patient waiting times. 
Therefore, it is important to take into account future patient arrivals while making scheduling decisions. 
Another challenge is the scheduling methodology employed at CHUM. Currently, scheduling tasks are done manually in an \textit{online} fashion, i.e. patients are scheduled one by one as they are admitted.
\textit{Online scheduling} has many disadvantages compared to \textit{batch scheduling}, where scheduling decisions are delayed to the end of the day or week, so that several patients can be scheduled at once. Batch scheduling can leverage accumulated information, and usually results in better schedules than online scheduling. 
However, it is difficult to adopt batch scheduling at CHUM. 
The scheduling staff need to call patients to confirm appointments and make adjustments when necessary. Online scheduling makes this communication easier. Therefore, it is required by CHUM to use online scheduling.%

We propose a novel \textit{online prediction-based approach} for \textit{dynamic radiotherapy scheduling}, which uses the historical arrival patterns of patients to inform scheduling decisions. 
An Integer Programming (IP) model is proposed to derive optimal \textit{offline schedules} for the variant of the problem where future arrivals are known in advance. 
Given a large number of instances with the same arrival rate and number of linacs, a regression model can be trained to detect the links between a patient's features, the (resource) allocation profile of the hospital, and the ideal schedule. The regression model can then predict a ``good'' waiting time for a patient. 
Patients are scheduled in an online fashion. 
Palliative patients are always scheduled on the  earliest possible date, while curative patients are scheduled using a prediction model that suggests a waiting time based on their treatment plan and the present occupancy of the linacs.  
In other words, the prediction model is trained to dynamically delay treatments of curative patients, to make space for palliative ones based on the current allocation profile of the hospital. 
This results in a robust scheduling approach that is able to take future arrivals into account. We refer to our proposed machine learning-based approach as the \textit{prediction-based approach}. 
We compare the prediction-based approach to myopic scheduling strategies, including the online greedy heuristic currently used at CHUM, a batch scheduling heuristic using a simple ordering rule, and two batch scheduling strategies using the IP model (daily and weekly).
The algorithms are tested in a simulation with a rolling horizon to evaluate their performance over a fixed-time period. 
Test datasets are generated using real data from CHUM. 
The prediction-based approach outperforms the other approaches for most problem settings, which vary by the number of linacs and the arrival rate. It results in the lowest average overdue time and average waiting time for palliative patients, while being comparable to other approaches with respect to average overdue times and average waiting times for curative patients. This demonstrates its ability to efficiently integrate information about future arrivals into daily scheduling decisions. 
Notably, the superiority of the prediction-based approach is more prominent in larger hospital settings with more linacs and more patients to treat (high arrival rate). 
The proposed approach also proves its robustness in a real problem instance, despite the high fluctuation in arrival rates observed in reality. Additionally, we demonstrate how our approach supports the explainability of the decision-making process using SHAP values. Explainability in decision-making enables better communication with clients. 
It is especially appreciated in the healthcare domain where most decisions are human-related. 
Explainability is a very active domain of research in Artificial Intelligence, yet receives little attention in the Operations Research community. 
As a hybrid method combining techniques from both domains, our approach supports explainable and interpretable scheduling decisions. 
	Our contributions can be summarized as follows:\looseness=-1

	\begin{itemize}
		\item We propose an online machine learning-based scheduling approach for radiation therapy. The machine learning model is trained on offline optimal solutions obtained via mathematical optimization to mimic the optimal policy without the online computational burden.\looseness=-1
		\item The proposed approach successfully reduces overdue treatments and waiting times for patients compared to other optimization-based approaches. We also successfully solve instances with up to 8 linacs, which represents a larger hospital size than those considered in previous studies. Furthermore, it is robust to the high fluctuation in arrival rates observed in practice.\looseness=-1
		\item The prediction-based approach is simple, easy to implement and maintain. It also offers solutions instantly, which makes it suitable for online scheduling problems.\looseness=-1
		\item We demonstrate how our approach supports explainability and interpretability in scheduling decisions, to our knowledge for the first time. This is highly appreciated in the healthcare domain.\looseness=-1 
	\end{itemize}

The rest of the paper is organized as follows. 
Section \ref{sec:problem_description} provides a problem description and literature review. Section \ref{sec:methodology} presents the methodology. Section \ref{sec:experiment} presents numerical results. Section \ref{sec:discussion} discusses theoretical and practical implications of our approach, and Section \ref{sec:conclusion} closes the paper with our conclusions.

\section{Problem description and related work}
\label{sec:problem_description}

In this section, we describe the radiotherapy scheduling problem arising at CHUM, and then present related work and situate our work in the literature.

\subsection{Problem description}

In RT treatment, a patient receives a series of radiation doses (fractions) delivered by a linac to the cancer site during  the treatment course, which typically consists of several consecutive days with breaks on the weekends. Multiple appointments are required. 
To simplify the problem, we assume that all linacs are identical. 
This does not affect the complexity of the problem, as special treatments that require special linacs are independent of the rest of the treatments, and can be scheduled separately in a sub-problem. 
In practice, each linac is associated with a technician team that will familiarize themselves with the technical settings of each patient at the beginning of their treatment. Therefore, hospital policy does not allow switching linacs during treatment. 
At CHUM, patients are divided into four categories with different waiting time targets based on their cancer type and condition. 
Palliative patients (categories P1 and P2) have their waiting time target set as 1 and 3 days, respectively. In curative patients (categories P3 and P4), the target is 14 and 28 days, respectively. 
To minimize the waiting time of patients, one only needs to assign the start date for treatment, as once a treatment starts it must be carried out daily. There are no treatments on weekends, but those gaps have been taken into account in the treatment protocols. The hospital does not allow gaps in treatment induced by planning. 
The waiting time of a patient is the time elapsed between their admission date, and the date of their first treatment. If a patient's treatment starts after the recommended deadline, the time delay is counted as overdue time.\looseness=-1

\begin{figure*}[]
\centering
\includegraphics[width=0.8\linewidth]{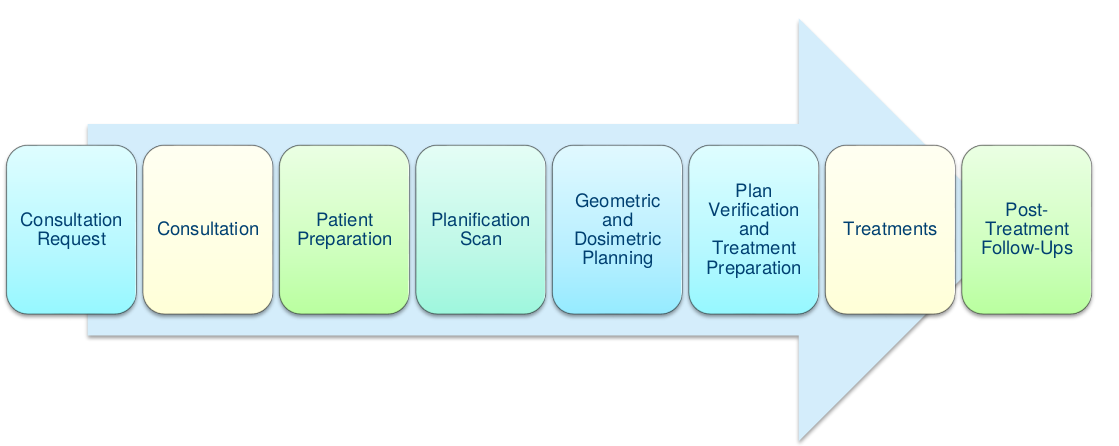}
\caption{Radiotherapy treatment workflow at CHUM\protect\footnotemark.}
\label{fig:CHUMprocess}
\end{figure*}

Cancer treatment is a complicated process with many procedures and parties involved. The treatment workflow used at CHUM is illustrated in Figure \ref{fig:CHUMprocess}. 
When a new consultation request is made, a patient first undergoes a consultation session where the doctor explains their condition and available options. If the patient agrees to proceed with the treatment, they will go through several preparation steps such as external consultation, exams, and a planification scan. 
Based on the patient's diagnosis, a treatment team then makes a personal treatment plan, which is verified and approved by a medical physicist before the treatments are prepared. 
The RT treatments are then carried out, with review and possible revision during the course of treatment. 
Finally, the treatment ends with post-treatments and follow-ups. Currently, the scheduling of treatments is performed manually at CHUM. 
Among various technical information, a treatment plan includes information about the number of fractions that a patient will receive, as well as the duration of each fraction.  
Since fraction durations at CHUM are always multiples of 5 minutes, we set the granularity of our schedules to 5-minute blocks. Each linac has a treatment capacity measured in time blocks.\looseness=-1   

\footnotetext{Figure provided by CHUM.}

\subsection{Literature review}
\label{sec:literature}
Patient scheduling has a long history and a rich literature. For a thorough review of the topic, we refer the readers to \citep{marynissen2019literature, cayirli2003outpatient,gupta2008appointment}. 
There are many ways to classify patient scheduling problems. 
To analyse related work, we adopt the classification of \cite{saure2012dynamic}, which divides patient scheduling into \textit{allocation scheduling} (or \textit{resource allocation}) and \textit{appointment scheduling}. Appointment scheduling assigns specific appointment times to patients on a given service day. In most appointment scheduling problems, the main concern is maximizing patient comfort, satisfying precedence constraints, dealing with uncertainty in appointment duration and cancellations, etc.
In contrast, allocation scheduling schedules patients in advance, taking medical capacity into account, without giving exact appointment times. Allocation scheduling usually involves patients of different types and urgency levels, therefore the main challenge is dealing with uncertainty, such as unknown future demands and cancellations. Algorithms for allocation scheduling are usually dynamic, meaning they focus on dynamically allocating medical resources on a rolling horizon. 
Our paper falls into the category of allocation scheduling. We review the literature on dynamic patient scheduling and radiotherapy scheduling. The literature is vast, so this not an extensive review, and only the most relevant papers are mentioned here.\looseness=-1

\subsubsection{Dynamic patient scheduling.}
There are different sources of uncertainty in healthcare: cancellations, uncertainty in service time, stochastic patient arrivals, deferrals in treatment plans, etc. While our paper focuses only on uncertainty resulting from stochastic patient arrivals, there are plenty of studies applying machine learning methods to tackle different sources of uncertainty. For example, \cite{srinivas2018optimizing} develop predictive models that identify the risk of a patient no-show and propose different scheduling rules that take into account the risk. A similar approach is used by \cite{salah2022predict} in an appointment scheduling system with the presence of stochastic factors from no-shows and uncertainty in service time.

Regarding uncertainty caused by stochastic patient arrivals, operating room (OR) planning has to deal with a high degree of uncertainty as ORs are usually shared between both elective (planned) procedures and emergency surgeries.  
\cite{gerchak1996reservation} propose a stochastic dynamic programming model for the OR planning problem, and then analyse the structure of the optimal policy.  
In the same problem, \cite{lamiri2008stochastic} propose a Monte Carlo optimization method that combines Monte Carlo simulation with Mixed Integer Programming (MIP), and show that their approach yields about a $4\%$ reduction of the overall cost compared to a deterministic model.\looseness=-1 

\cite{vermeulen2009adaptive} present an adaptive approach to schedule CT scans at a radiology department. 
In their approach, the allocation of capacity to different patient groups is flexible and adaptive to the current and expected future situation. However, the algorithm is specific to their case study. 
\cite{patrick2008dynamic} model a similar problem as a Markov decision process (MDP), and solve it using simulation-based approximate dynamic programming (ADP). \cite{saure2012dynamic} and \cite{saure2020dynamic} extend this approach to more complicated problems.
\cite{schuetz2012approximate} investigate a capacity allocation problem in the service industry with requests of different priorities. The authors also adopt a similar approach,  combined with a discrete event simulation.\looseness=-1

In another problem family, patient admission scheduling (PAS), \cite{ceschia2012modeling} propose a MIP model and Simulated Annealing heuristic to solve a PAS problem under uncertainty. 
They propose a \textit{degree of dynamism}, adopted from dynamic vehicle routing problems, to denote a level of urgency. 
 \cite{ceschia2016dynamic} then extend the problem with more realistic constraints and propose a local search-based approach.
\cite{zhu2019compatibility} study the compatibility of short and long-term objectives for the dynamic PAS. 
The authors propose a short-term MIP model which adjusts the objective functions based on patient delay or idle resources.\looseness=-1 

\subsubsection{Radiotherapy scheduling.}

As opposed to patient scheduling, RT scheduling is a relatively new field of research. 
A review of RT scheduling can be found in \citep{kapamara2006review, vieira2016operations}. 
The first two mathematical models for RT scheduling are proposed by \cite{conforti2008optimization} to maximize the number of treated patients within a given horizon with identical treatment times and a single linac. The model assigns patients to treatment slots of equal length, which is referred to as \textit{block} scheduling. 
The authors then extend their models in \citep{conforti2010non} to allow for more realistic settings where treatment duration can vary in length (\textit{non-block scheduling}). 
\cite{petrovic2006algorithms} and \cite{petrovic2008constructive} propose several constructive heuristics based on different prioritised rules, and improve the solutions using metaheuristics. They use block scheduling where treatment on a linac is of fixed duration, and depends on the energy type of the linac. 
\cite{burke2011integer} propose a non-block IP model to assign patients to days and linacs, and perform several simulations to investigate whether delaying the scheduling decision can lead to better schedules. 
Heuristic methods are also used by \cite{kapamara2009heuristics} to schedule patients in both pre-treatment and treatment stages. 
Pre-treatment for RT patients is also considered by \cite{castro2012combined}. The authors solve a multi-objective problem by formulating it as a series of single-objective scheduling problems.\looseness=-1   

Another form of RT treatment is particle therapy, which uses a centralized machine to deliver an ion beam to different treatment rooms. Most studies in particle therapy focus on assigning specific appointment times for treatments to optimize the beam's utilization, i.e., minimizing the idle time of the machine. \cite{maschler2016particle} propose an exact model, which is proven to be highly intractable. As a result, metaheuristics are widely used for particle therapy treatment scheduling \citep{maschler2016particle,maschler2018particle,vogl2019scheduling}. 
In another context, \cite{vieira2020radiotherapy} point out that most cancer centers in Europe are capable of treating their patients within the recommended waiting period. Therefore, they focus on meeting patients' preferences for treatment times. 
\cite{frimodig2019models} propose an IP and two constraint programming (CP) models for scheduling RT treatments. Their generated instances include an expected number of future arrivals, which they use to predict future resource utilization.\looseness=-1 

All of the aforementioned methods are tested on a static horizon without taking future arrivals into account. RT scheduling involves patients of different priorities, so using a dynamic resource allocation scheme that takes uncertainty of treatment demand into account is highly important. 
\cite{pham2021two} propose a two-phase approach for RT scheduling, and implement a simulation to evaluate the algorithm dynamically on a rolling horizon using different scheduling strategies. 
However, the approach is still myopic, i.e., it does not take future arrivals into account when making scheduling decisions. 
To handle high-priority patients, many cancer centers reserve a number of treatment slots per day for emergency treatments. This approach is used in several papers \citep{petrovic2008constructive,legrain2015online,pham2021two,frimodig2019models}. Nonetheless, it is difficult to set a proper threshold.\looseness=-1  

To the best of our knowledge, there are only two approaches for dynamic RT scheduling which take future events into account. 
The first approach is MDP, used by \cite{saure2012dynamic} to provide a dynamic policy for RT scheduling. The problem is modeled as a discounted infinite-horizon MDP, and the equivalent Integer Programming (IP) model is solved using column generation. 
The same approach is also used by \cite{gocgun2018simulation}, but they also allow for the cancellation of treatments.
In contrast, \cite{legrain2015online} propose a hybrid method combining stochastic optimization and online optimization to solve the problem in an online fashion. 
Stochastic programming and MDPs are both prone to scaling problems which makes it difficult to apply either approach in large hospitals. 
The example studied by \cite{saure2012dynamic} has an arrival rate of 8.25 requests per day with 120 appointment slots, which is equivalent to three linacs. 
	\cite{legrain2015online} consider a 20-minute time slot for each patient, and test the algorithm on instances with up to two linacs and an arrival rate less than $3.5$ requests per day.
	In addition, they are algorithmically heavy, and may be difficult to implement and maintain in real-world applications. In the literature, there are few attempts that tackle the scaling problem resulting from using stochastic programming with machine learning techniques. One example can be found in the context of two-stage stochastic programming \citep{larsen2021predicting}, where the authors use supervised machine learning to predict the computationally expensive second stage and apply the algorithm to a railway demand and capacity management problem. 
Our approach aims to fill this gap in the RT scheduling literature. We propose an algorithm that dynamically allocates resources based on expected future arrivals, and targets larger instances than those considered before in the literature, i.e., up to 8 linacs.\looseness=-1

\section{Methodology}
\label{sec:methodology}
Section \ref{sec:mathmodel} presents the mathematical model, and Section \ref{sec:scheduling_strategies} presents several scheduling strategies. 
 The remainder of Section \ref{sec:methodology} explains the prediction-based approach.\looseness=-1 

\subsection{Mathematical model}
\label{sec:mathmodel}
The problem being solved consists of finding the best schedule of treatments for a set of patients $\mathcal{P}$, given a set of linacs $\mathcal{L}$. 
The sets of palliative patients and curative patients are denoted as $\mathcal{P^P}$ and $\mathcal{P}^{C}$, respectively (so $\mathcal{P} = \mathcal{P^P} \cup \mathcal{P}^{C}$). Each instance consists of a set of fixed patients with appointments made from the previous scheduling decisions, denoted as $\bar{\mathcal{P}}$, while the set of new patients is denoted as $\hat{\mathcal{P}}$ (so $\mathcal{P} = \bar{\mathcal{P}} \cup \hat{\mathcal{P}}$). 
The set of working days in the planning horizon is denoted as $\mathcal{T}$. 
Each patient $i \in \mathcal{P}$ has an admission date $a_i$, a ready date $r_i$ which is the earliest date the patient can begin treatment, and a due date $d_i$ which is the recommended deadline to start the treatment. 
Each patient $i$ has a personal treatment plan specifying the number of fractions $I_i$, 
and the duration of each fraction ($p_i$). 
Each linac $l$ has a capacity ($C^t_l$) on day $t$, measured in blocks of 5 minutes. 
$\hat{C}^t_l$ represents the available capacity of linac $l$ on day $t$ after deducting the fixed appointments from the previous scheduling decisions.\looseness=-1

\begin{table}[]
	\small
	\centering
	\begin{tabular}{cl|cl}
		Notation & Explanation  & Notation & Explanation \\ \hline
	$\mathcal{L}$	& set of linacs 				& $a_i$ & admission date of patient $i$\\
	$\mathcal{P^P}$	&  set of palliative patients  & $\alpha_i$ & admission time of patient $i$ (consists of date and time)\\ 
	$\mathcal{P}^{C}$	& set of curative patients & $r_i$ & ready date of patient $i$ ($r_i \geq a_i$) \\ 
	$\bar{\mathcal{P}}$	&  set of fixed patients  & $d_i$ & due date of patient $i$ ($d_i \leq r_i$)\\ 
	$\hat{\mathcal{P}}$ & set of new patients & $I_i$ & number of fractions of patient $i$ \\
	$\mathcal{T}$ & set of working days in the planning horizon & $p_i$ & duration of each fraction of patient $i$ \\
	$C^t_l$ & total capacity of linac $l$ on day $t$ &$\hat{C}^t_l$ & available capacity of linac $l$ on day $t$ 
	\end{tabular}
\caption{List of notation.}
\end{table}

Our objective is to minimize overdue time and waiting time by determining the assignment of patients' first fractions to dates and linacs. Once treatment starts, it is carried out every day until the required number of fractions is achieved. Subsequent fractions of each patient should be carried out on the same linac as the first treatment. 
A portion of linac capacity is reserved for palliative patients. Parameter $\gamma$ ($\gamma < 1$) indicates the percentage of reserved capacity.
We define a binary variable $x^i_{tl}$ on the set of new patients $\hat{\mathcal{P}}$, that holds value $1$ if patient $i$ receives their first treatment on day $t$ linac $l$. The model is as follows:\looseness=-1

\vspace*{-0.5cm}

\begin{align}
\mbox{minimize } &
\sum_{i \in \hat{\mathcal{P}}} \sum_{t \in \mathcal{T},  t > r_i} \sum_{l \in \mathcal{L}} \omega_1 (t-a_i)log(t-a_i+1) x^i_{tl}			\nonumber	\\
& + \sum_{i \in \hat{\mathcal{P}}} \sum_{t \in \mathcal{T}, t > d_i} \sum_{l \in \mathcal{L}}  \omega_2  (t-d_i)log(t-d_i+1) x^i_{tl}		\label{eq:20}
\end{align}

\vspace*{-0.2cm}
Subject to
	\begin{align}
	&\sum_{t \in \mathcal{T}} \sum_{l \in \mathcal{L}} x^i_{tl} = 1 	&	\forall i \in \hat{\mathcal{P}} \label{eq:30}	\\
	&x^i_{tl} = 0 	& \forall i \in \hat{\mathcal{P}}, l \in \mathcal{L}, t \in \{0, \ldots, r_i-1\} \label{eq:31}				\\
	&\sum_{i \in \hat{\mathcal{P}}} \sum_{t' = max\{0,t - I_i + 1\}}^t p_i x^i_{t'l} \leq \hat{C}^t_l  	& \forall t \in \mathcal{T},  l \in \mathcal{L} \label{eq:32}			\\
	&\sum_{i \in \mathcal{P}^{\mathcal{C}}} \sum_{t' \in \{t - I_i + 1, \ldots, t\}} p_i x^i_{t'l} \leq max\{0, \hat{C}^t_l - \gamma C^t_l\} 		  
	&  \forall t \in \mathcal{T},  l \in \mathcal{L} 	\label{eq:33}	\\
	& x^i_{tl} \in \{0,1\}  & \forall i \in \hat{\mathcal{P}}, t \in \mathcal{T}, l \in \mathcal{L} \label{eq:34}
	\end{align}

The objective function (\ref{eq:20}) minimizes waiting time (the first term) and overdue time (the second term) with the respective weights $\omega_1$ and $\omega_2$. 
By setting $\omega_2 \gg \omega_1$, we heavily penalize overdue time over waiting time. 
The log function is a non-linear function utilized to enforce an equal distribution of waiting time and overdue time between patients. 
Constraints (\ref{eq:30}) ensure all patients are scheduled within the planning horizon. 
Constraints (\ref{eq:31}) ensure no patients are scheduled before their ready date. 
Constraints (\ref{eq:32}) make sure linac capacity is respected for the entire course of treatment. 
Constraints (\ref{eq:33}) reserve a portion of linac capacity for palliative patients.\looseness=-1

\subsection{Scheduling strategies}
\label{sec:scheduling_strategies}
In this section, we propose several scheduling strategies for radiotherapy treatments. All strategies utilize online scheduling for palliative patients, i.e., their treatments are scheduled for the earliest possible date on an available linac when the treatment request is submitted.  
A portion of linac capacity is reserved for palliative patients. 
The proposed strategies differ in the manner in which curative patients are handled. 
All scheduling strategies are summarized in Table \ref{tab:schedulingstrategy}.\looseness=-1 

\begin{table}[h!]
\centering
\small
\begin{tabular}{c|c|c}
	Strategy & Scheduling palliative patients & Scheduling curative patients \\	\hline
	offline  & one time & one time \\
	online-greedy  & at admission & at admission \\
	daily-greedy  & at admission & every day \\
	daily-IP  & every day & every day \\
	weekly-IP  & every day & every Friday \\
	prediction-based  & at admission & at admission
\end{tabular}
\caption{Scheduling strategies.} 
\label{tab:schedulingstrategy}
\end{table}

\subsubsection{Online scheduling with a greedy heuristic}
\label{sec:greedy}

\begin{figure}[ht]
\centering
\includegraphics[width=0.9\linewidth,height=0.3\linewidth]{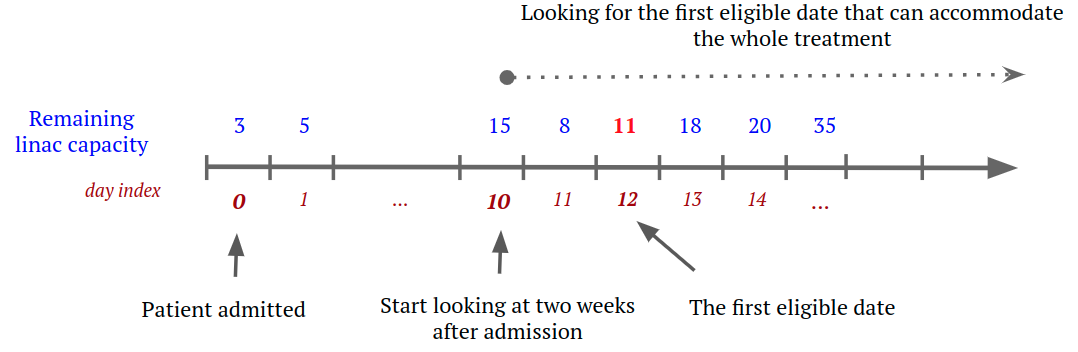}
\caption{The greedy heuristic for an instance with one linac. A curative patient (category P4) with 3 fractions, each of which requires 10 time slots, is admitted on day 0. The algorithm looks forward two weeks, starting from day 10 (only business days are indexed). The patient is scheduled on day 12, the first day with enough capacity in subsequent days to accommodate the whole treatment.}
\label{fig:greedy}
\end{figure}

This strategy is the closest one to the approach used by scheduling clerks at CHUM. Once a patient is admitted (usually when their treatment plan is approved), the algorithm looks up the following one or two weeks depending on the patient's category (P3 or P4). The first date that can accommodate all of the patient's fractions is then chosen. A date is considered eligible if on this day, and the following $(I_i - 1)$ days, there is a linac with enough remaining capacity to treat the patient, where $I_i$ represents the number of fractions assigned to the patient. 
An illustration of the greedy heuristic can be found in Figure \ref{fig:greedy}.\looseness=-1

\subsubsection{Batch scheduling}
When batch scheduling, treatment requests are accumulated for a period of time, and then a scheduling decision is made by the end of the period, e.g., daily or weekly. Each batch of scheduling decisions includes a set of patients admitted the day after the previous scheduling decision was made.  
We propose two strategies to schedule patients in batches: greedy-based, and IP-based. The greedy-based heuristic sorts patients in descending order by their due day (which is determined by their priorities), numbers of fractions, and fraction lengths. Patients are then scheduled in a greedy manner following the given order. This prioritizes patients with more fractions and longer fraction lengths, as it is more difficult to find an eligible start day for them. In IP-based batch scheduling, patients are scheduled using the IP model in Section \ref{sec:mathmodel}.\looseness=-1

\subsubsection{Offline scheduling}
\label{sec:static}
In offline scheduling, we assume that all patient arrivals are known in advance. 
First, the algorithm schedules palliative patients for the earliest eligible day after their arrival. Curative patients admitted during the simulation period are then scheduled using the IP model in Section \ref{sec:mathmodel}, without reserving a portion of linac capacity for palliative patients. 
The scheduling decision is made on the first day of the scheduling period. 
Having all future patient arrivals known in advance is an unrealistic assumption, but it provides a bound for other scheduling strategies. No strategy can bypass offline scheduling due to the lack of information on future arrivals. It also provides a picture of how a ``perfect schedule'' would look if we had a precise prediction of future arrivals.\looseness=-1

\subsubsection{Online scheduling using regression - a machine learning approach}
\label{sec:reg_approach}
\begin{figure*}[]
\centering
\includegraphics[width=0.85\linewidth, height=0.28\linewidth]{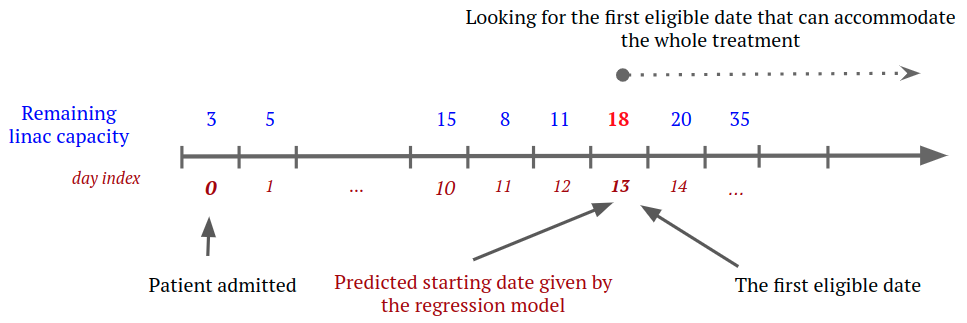}
\caption{Prediction-based scheduling for the same instance as Figure \ref{fig:greedy}. The algorithm starts the search from the predicted starting day (day 13). The patient is scheduled to the first eligible day (day 13).}
\label{fig:regression}
\end{figure*}

This scheduling strategy is the main contribution of the paper. Scheduling is carried out in an online fashion similar to the greedy heuristic. However, instead of looking forward one or two weeks, we apply a regression model to predict a start date for each patient. The algorithm then looks for an eligible date starting from the predicted day. 
An illustration of the algorithm is given in Figure \ref{fig:regression}. The details of the regression model will be presented in Section \ref{sec:MLapproach}.\looseness=-1 

\subsection{Prediction-based approach - a machine learning-based algorithm}
\label{sec:MLapproach}
In most cancer centers, the number of linacs is fixed. The arrival rate of patients is usually stable and can be modelled as a Poisson distribution. 
Given an instance with a number of linacs and an arrival rate, an optimal offline solution (referred to simply as an \textit{offline solution}) given by offline scheduling (Section \ref{sec:static}) provides an ``ideal'' schedule where all future arrivals are known in advance.
All other scheduling strategies based on the gradual realization of patient arrivals share the same goal of achieving solutions as close as possible to offline solutions. 
The motivation for our approach is the idea that given a large enough number of instances and their offline solutions, we might be able to learn the patterns in the ideal optimal waiting time of patients. Thus, given the present occupancy of linacs and the treatment plan of a patient, we can predict a ``good'' waiting time for the patient. 
In the following section, we describe our method of training the regression model and applying it in a machine learning-based scheduling algorithm, which we refer to as a \textit{prediction-based approach}.\looseness=-1

\subsection{Training the regression model}
\label{sec:train_regmodel}
A problem instance is characterized by a number of linacs $|\mathcal{L}|$ and an arrival rate $\lambda$. 
A pair $(|\mathcal{L}|, \lambda)$ is referred to as an \textit{instance setting}, which represents a typical cancer center with a fixed number of linacs and a stable arrival rate. 
In all our instances, all linacs have a capacity of 120 blocks of 5 minutes a day. 
As each hospital has its own historical ideal scheduling patterns, an instance setting is associated with a separate regression model. 
For each instance in the training set, an offline solution is obtained by solving the offline scheduling as in Section \ref{sec:static}. A set of training examples is then constructed from the instance and its offline solution.\looseness=-1 

To explain how training examples are constructed, we first define some terms. 
A problem instance has a set of simulation days (denoted as $\mathcal{K}$), each of which observes a set of patients admitted to the center (denoted as $\mathcal{P}_d, d \in \mathcal{K}$). A problem instance consists of the daily flows of patients admitted to the hospital during the simulation period, which is referred to as a patient flow $\zeta$, where $\zeta = \{\mathcal{P}_d, \forall d \in \mathcal{K}\}$.  
A solution to an instance is the assignment of patients to a set of dates ($\mathcal{D}$) and linacs ($\mathcal{L}$): 
$s: \mathcal{P} \times \mathcal{D} \times \mathcal{L} \rightarrow \{0,1\}$, 
where $s^i_{dl} = 1$ if patient $i$ is assigned to day $d$ linac $l$.\looseness=-1 

The \textit{present time point} at any moment during the simulation period is denoted as $\phi$, and the corresponding date is denoted as $d(\phi)$. 
At time point $\phi$, the \textit{present capacity} of linacs on date $d$ is the available capacity after removing the treatment slots occupied by patients admitted (and scheduled) before the time point $\phi$ on the corresponding date. 
$
\hat{c}^\phi_d = C - \sum_{i \in \mathcal{P}, l \in \mathcal{L} | s^i_{dl} = 1, \alpha_i < \phi} {p_i},
$
where $C$ is the total linac capacity, and $p_i$ and $\alpha_i$ are the fraction duration and admission time point of patient $i$, respectively. Note that $\alpha_i$ is the admission time point, which includes both admission date $a_i$ and admission time. We distinguish between $\alpha_i$ and $a_i$, because the present capacity $\hat{C}^\phi$ varies by time points during the day. 
The \textit{state} of an instance at time point $\phi$, denoted as  $\hat{C}^\phi$, is the set of \textit{present capacity} of all days in the \textit{sample horizon}, 
$
\hat{C}^\phi = \{\hat{c}^\phi_d, \forall d \in \mathcal{D}^\phi \}, 
$
where $\mathcal{D}^\phi$ is the set of days in the sampling horizon at the time point $\phi$. In our problem, the sampling horizon is set as 50 days after $d(\phi)$, $\mathcal{D}^\phi = \{ d(\phi), d(\phi) + 1, \ldots, d(\phi) + 49 \}$, 
since in our realistic instance settings it is unlikely that a patient is not scheduled within 50 days of their admission. Therefore, when making a scheduling decision, the linac capacity search is restricted to a window of 50 days after their admission.\looseness=-1

Given an instance and its offline solution $s^*$, a training example consisting of an input vector and a label is created for each curative patient in the patient flow. 
The label of patient $i$ is the patient's waiting time $w_i$ in the offline solution. 
The input vector consists of the \textit{state} measured at the patient's admission time $\phi = \alpha_i$, and a vector of the patient's features. 
We select the features that are the most relevant to the scheduling task, i.e., the patient's ready date $r_i$, due date $d_i$, number of sessions $I_i$, and fraction length $p_i$.
Thus, a training example for patient $i$ is a tuple $(X_i, w_i)$, with input vector $X_i = \{ \hat{C}^{\phi, \phi = \alpha_i}, r_i, I_i, d_i, p_i\} $.{The collinearity of the features are tested in Appendix \ref{app:trainingresult}.
\looseness=-1

The algorithm for creating training examples is presented in Algorithm \ref{alg:gen_trainingexamples}. 
As the regression model is used to schedule curative patients only, one training example is created for each new curative patient in the instance. 
At the beginning of the algorithm, the present time point is set as the beginning of the first day in the simulation period $\mathcal{K}$, and the present capacity $\hat{C}^\phi$ is calculated accordingly. 
For each day $d \in \mathcal{K}$, the set of patients admitted on that day ($\mathcal{P}_d$) is iterated over in chronological order. If a patient is curative, a training example is created and added to the set of training examples. After each iteration, regardless of the patient's priority, the present capacity $\hat{C}^\phi$  is updated with the appointments of the corresponding patient from the offline solution $s^*$. The process continues until all patients in the patient flow are visited. 
Different regression models were tested (Appendix \ref{app:trainingresult}), and XGBoost was chosen due to the low training time and high precision. \looseness=-1 

\begin{algorithm}[]
\small
\SetAlgoLined
\SetKwInOut{Input}{input}
\SetKwInOut{Output}{output}
\Input{instance $ins$, its offline solution $s^*$}
\Output{set of training examples $\mathcal{E}$}
$\mathcal{E} \leftarrow \{\}$  \;
$\phi \leftarrow$ the beginning of the first day in the simulation period $\mathcal{K}$	\;
calculate $\hat{C}^\phi$	\;
\ForEach{day $d$ in the simulation period $\mathcal{K}$}{
	\ForEach{patient $i$ in $\mathcal{P}_d$}{
		\If{$i$ is curative}{
			$X_i \leftarrow \{ \hat{C}^{\phi}, r_i, I_i, d_i, p_i\} $ \;
			$w_i \leftarrow$ waiting time of $i$ in $s^*$ \;
			$\mathcal{E} \leftarrow \mathcal{E} \cup (X_i ,w_i)$ \;
		}
		$\phi \leftarrow \alpha_i$	\;
		update $\hat{C}^\phi$ with appointments of patient $i$ \; 
	}
}
\caption{Generating training examples}
\label{alg:gen_trainingexamples}
\end{algorithm}

\subsection{Using the regression model}
The regression model is embedded in the prediction-based algorithm (Algorithm \ref{alg:regression_approach}) to schedule RT treatments. 
When a new patient is admitted, there are two possibilities. If the patient is palliative, they will be scheduled on the first eligible date. Otherwise, an input vector is constructed as described in Section \ref{sec:train_regmodel}. 
The input vector is fed to the trained regression model to get a predicted waiting time, from which a predicted starting date is derived and utilized in the prediction-based algorithm to construct a schedule. 

\begin{algorithm}[]
\small
\SetAlgoLined
\SetKwInOut{Input}{input}
\SetKwInOut{Output}{output}
\Input{instance $ins$, a trained regression model $\mathcal{M}$}
\Output{schedule $s$}
Initialize an empty schedule $s$  \;
$\phi \leftarrow$ the beginning of the first day in the simulation period $\mathcal{K}$	\;
calculate $\hat{C}^\phi$	\;
\ForEach{day $d \in \mathcal{K}$}{
	\ForEach{patient $i$ in $\mathcal{P}_d$}{
		\If{$i$ is palliative}{
			$d^* \leftarrow $ the first eligible date for $i$ starting from $max ( d, r_i )$ \;
		}
		\If{$i$ is curative}{
			$X_i \leftarrow \{ \hat{C}^{\phi}, d, r_i, I_i, d_i, p_i\} $ \;
			$\bar{w} \leftarrow$ predicted waiting time given by $\mathcal{M}(X_i)$ \;
			$d^* \leftarrow $ the first eligible starting date for $i$ from $max(d + \bar{w}, r_i)$ \;
		}
		$s \leftarrow $ update the schedule by scheduling all fractions of $i$ with the starting date $d^*$ \;
		$\phi \leftarrow \alpha_i$	\;
		update $\hat{C}^\phi$ with new appointments of patient $i$ \; 
		
	}
}
\caption{Prediction-based scheduling}
\label{alg:regression_approach}
\end{algorithm}

\section{Numerical results}
\label{sec:experiment}
Three experiments were carried out. 
The first experiment (Section \ref{sec:tuning}) examines the behaviour of the proposed approaches at different reservation rates. The second experiment (Section \ref{sec:exp_result_generateddata}) evaluates the algorithms on generated data in different scenarios. Finally, the third experiment (Section \ref{sec:exp_realins}) shows the results on real patient flow at CHUM. 
For each instance setting, 500 instances were generated, 400 of which were used for training the regression model, while the remaining 100 were used for testing. 
We compare the six scheduling strategies listed in Section \ref{sec:scheduling_strategies}. 
All of the algorithms are implemented in Python. We used CPLEX 20.1 as the MIP solver. The experiments were run on a Linux-based PC cluster. To generate offline solutions, each instance was solved using a single thread with a budget of 15 Gb of RAM, and a time budget of 15 hours.\looseness=-1   

\subsection{Data overview}
\label{sec:data_overview}

We have access to historical data from CHUM for a 2-year period, from September 2017 to July 2019, which includes 4,538 patients. 
CHUM is equipped with 12 treatment rooms, 5 of which contain identical generic linacs. 
4 rooms are dedicated to special treatments that must be bound to those specific rooms. 
The remaining 3 rooms can be used for generic treatments, as well as a small percentage of specialized treatments. 
Patients are classified into four categories with different treatment deadlines. The majority of patients are curative (more than $70\%$) with 14-day or 28-day treatment deadlines. 
The remaining patients are mostly palliative type P2 with 3-day treatment deadlines. A small portion of patients (type P1) requires urgent treatment with a 1-day deadline. In the given period, more than $70\%$ of P2 and P3 patients at CHUM did not meet their recommended treatment deadline. 
Patient categories along with their waiting time targets, percentage of overdue treatments, and average waiting time, are listed in Table \ref{tab:categories}. 
The fraction length of patients at CHUM ranges from 10 to 165 minutes, with the majority of patients having 25-minute fractions (more than $50\%$). 
The number of fractions varies from 1 to up to 45 sessions. 
Most palliative patients have less than 5 fractions, while the number of fractions of curative patients ranges from 1 to 35 sessions, with peaks at 5, 15, 20, 25, 30, and 33 sessions.\looseness=-1 

\begin{table*}[]
\small
\centering
	\begin{tabular}{c|c|c|c|c}
		Category &  Proportion (\%) &  \begin{tabular}[c]{@{}c@{}}Treatment deadline \\ (days)\end{tabular} & \begin{tabular}[c]{@{}c@{}}Percentage of\\ overdue treatment (\%)\end{tabular} & \begin{tabular}[c]{@{}c@{}}Average\\ waiting time (days)\end{tabular} \\ \hline
		P1  & 0.44     & 1  & 14.29 & 1.09 \\ 
		P2  & 27.14     & 3   & 79.89 & 6.91 \\ 
		P3  & 41.36     & 14  & 74.55 & 18.11 \\ 
		P4  & 31.06     & 28  & 29.89 & 22.59 \\ 
	\end{tabular}%
\caption{Overdue treatment and average waiting times for cancer patients at CHUM in 2017-2018. }
\label{tab:categories}
\end{table*}

\subsection{Data generation}
\label{sec:exp_datagen}

Before describing the process of generating our instances from CHUM's real data, we introduce some technical terms. 
A \textit{personal treatment plan}, which provides information such as the number of fractions and fraction length, is tailored for an individual patient by a team of physicians and specialists based on the patient's diagnosis and condition. 
 For convenience, we use the term \textit{treatment plan} to refer to a \textit{personal treatment plan}. 
A \textit{patient pool} consists of all historical treatment plans taken from CHUM's dataset. 
A problem instance consists of an \textit{instance scenario} and a \textit{patient flow}. 
An instance scenario specifies a number of linacs and their capacity, along with a set of appointments made during the previous batch of scheduling decisions. 
A patient flow is the daily flow of new patients admitted to the hospital during a \textit{simulation period}. 
Given an arrival rate $\lambda$ and a set of simulation days $\mathcal{K}$, a set of $|\mathcal{K}|$ numbers is generated using a Poisson distribution with an event rate of $\lambda$. 
For each day $d \in \mathcal{K}$, a set of virtual patients is generated by randomly selecting a corresponding number of treatment plans from the patient pool. 
Then a ready date is generated for each patient based on their category. The ready date for P1 patients is always the same as their admission date. P2 patients have ready dates ranging from 0 days after their admission date to 2 days after their admission date, while curative patients (P3 and P4) have ready dates ranging from 5 to 7 days after their admission date. 
The due date is calculated from the admission date and the patient’s category listed in Table \ref{tab:categories}.\looseness=-1 

An instance scenario is generated as follows. First, we generate an instance flow, which then serves as an input for ``warm-up'' simulation using the greedy heuristic (Section \ref{sec:greedy}) to fill up the linacs, starting from an empty schedule. 
The warm-up simulation stops when the occupancy of the linacs reaches $90\%$ of its capacity on any day. The $90\%$ threshold is based on the real schedules at CHUM. The following day is usually almost full, with a couple of available slots that normally account for no more than $10\%$ of the total capacity. 
All appointments on the dates after the end of the warm-up period are used as the new instance scenario. An illustration is shown in Figure \ref{fig:generator}.\looseness=-1  

\begin{figure}[h!]
\centering
\includegraphics[width=0.7\linewidth]{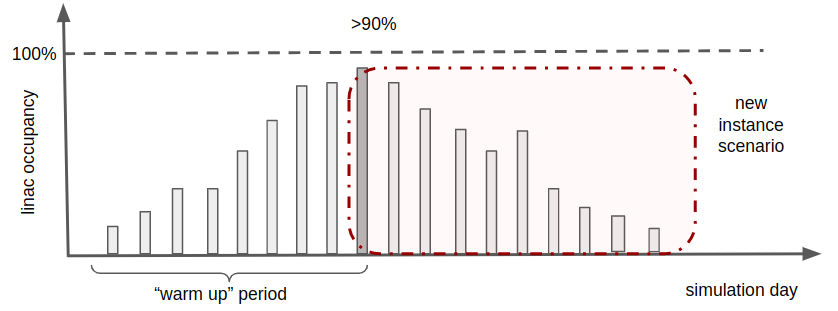}
\caption{Illustration on how an instance scenario is generated.}
\label{fig:generator}
\end{figure}

\subsection{Tuning the reservation rates for palliative patients}
\label{sec:tuning}
The reservation rate plays a different role in the prediction-based approach than it does in myopic approaches. 
In myopic approaches, the reserved capacity makes space for future arrivals of palliative patients. A low reservation rate might lead to a shortage of treatment slots for urgent cases. 
In contrast, a high reservation rate does not fully exploit the treatment facility, which leads to delayed treatments for curative patients.  
In the prediction-based approach, an implicit ``dynamic reservation'' is applied by delaying the treatments of curative patients dynamically to make space for palliative ones based on the present allocation profile. 
Hence, the reserved capacity plays the role of a ``safe corridor'' to absorb the effect of unexpected high arrival. 
The goals of the experiment in this section are twofold: (1) to examine the behaviour of approaches with different reservation rates; and (2) to identify the reservation rate that yields the best performance for all approaches.\looseness=-1  

The experiment was carried out using a setting with 4 linacs and an arrival rate of 6.0. 100 instances were tested with five different reservation rates: $0\%$, $5\%, 10\%, 15\%$, and $20\%$. As there is no reservation in offline scheduling, five scheduling strategies were tested: online greedy heuristic, daily greedy heuristic, daily-IP, weekly-IP, and prediction-based. The results are shown in Figure \ref{fig:ratetuning}. 
We can see from the figure that a higher reservation rate is indicative of a lower overdue time for palliative patients. However, the trade-off that comes with a low overdue rate, is a high number of overdue times for curative patients. 
Among all tested reservation rates, the prediction-based approach always gives lower overdue times in palliative patients by a large magnitude. Consider a reservation rate of $10\%$; other myopic approaches result in around $20$ days of overdue time on average in palliative patients, while the prediction-based approach keeps that metric below 2 days. 
The prediction-based approach yields a slightly higher average overdue time for P3 compared to other approaches, while maintaining similar values for P4. A similar profile can be seen for other reservation rates. 
Nevertheless, one can observe that as the reservation rate increases, the overdue times for curative patients increase significantly when using myopic approaches, while the decrease for palliative patients is not proportional. 
From the aforementioned observations, we conclude that the prediction-based approach offers a dynamic reservation that outperforms the flat reservation policy of other myopic approaches, regardless of the reservation rate used. 
After discussing our results with a consultant from CHUM, we decided that a reservation rate of $10\%$ results in the best trade-off between the delays in curative and palliative patients. 
This choice is further supported by the data showing that the total treatment duration of palliative patients accounts for around $8\%$ of the total treatment duration for all patients. 
Therefore, we adopt a reservation rate of $10\%$ for all scheduling approaches.\looseness=-1 

\begin{figure}
	\centering
	\includegraphics[width=\linewidth]{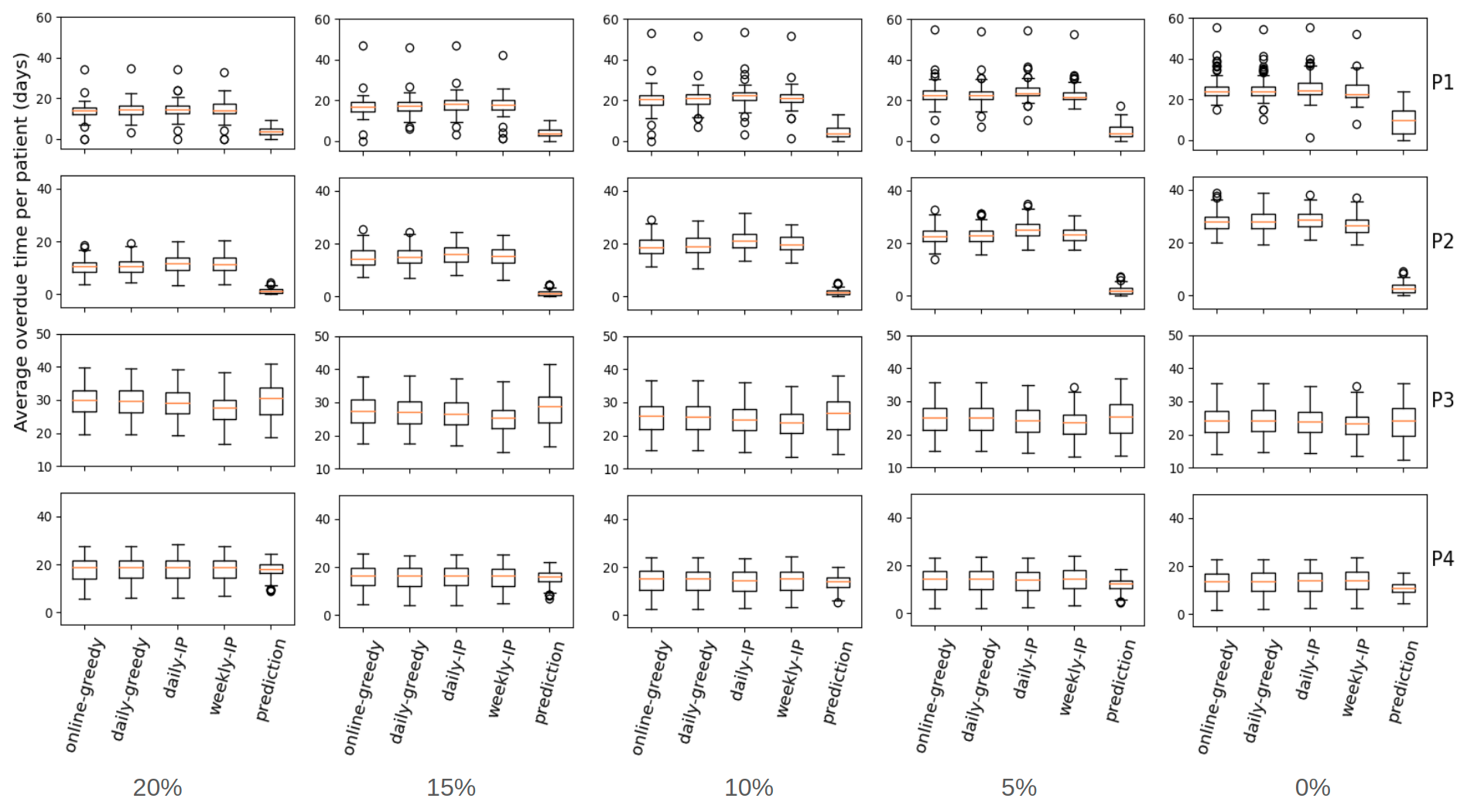}
	\caption{Average overdue time with different reservation rates - 4 linacs, arrival rate of 6.0.}
	\label{fig:ratetuning}
\end{figure}
\color{black}

\subsection{Numerical results on generated instances}
\label{sec:exp_result_generateddata}
To examine the algorithms' performance on different hospital sizes and crowding levels, we generated instances with a wide range of linac numbers and arrival rates. 
First, we identified a realistic range of arrival rates for a hospital with a given number of linacs using two types of \textit{capacity simulation}. 
For the first capacity simulation, we ran the greedy algorithm in Section \ref{sec:greedy} without imposing the capacity constraints. 
We then measured the average weekly occupancy rate. Figure \ref{fig:4linacs}a shows the results of the first capacity simulation for 4 linacs with arrival rates of 5.0, 5.5, 6.0, and 6.5, respectively. The red dotted line represents the linac capacity. Arrival rates 6.0 and 6.5 result in overload for multiple weeks during the simulation period. An arrival rate of 5.0 is in the safe zone, with the occupancy rate remaining mostly below the linac capacity. However, an arrival rate of 5.5 can cause overload in some weeks, but generally stays within the safe zone. 

For the second capacity simulation, we ran the greedy algorithm with capacity constraints and then observed the progression of the weekly average waiting time. If the average waiting time increased over time, the arrival rate was too high for a hospital of the given size to handle. A stable average waiting time indicates that the arrival rate is reasonable. Figure \ref{fig:4linacs}b shows the results of the second capacity simulation for 4 linacs. Observe from the figure that an arrival rate of 6.5 causes the average waiting time to increase rapidly over time. This matches the observation from the first capacity simulation. 
The remaining arrival rates result in stable lines, with some peaks in the arrival rates of 5.5 and 6.0. 
The two capacity simulations complement each other, and we can conclude that arrival rates from 5.0 to 6.0 are reasonable for a hospital with 4 linacs. 

\begin{figure}[h]
	\centering
	\includegraphics[width=\linewidth]{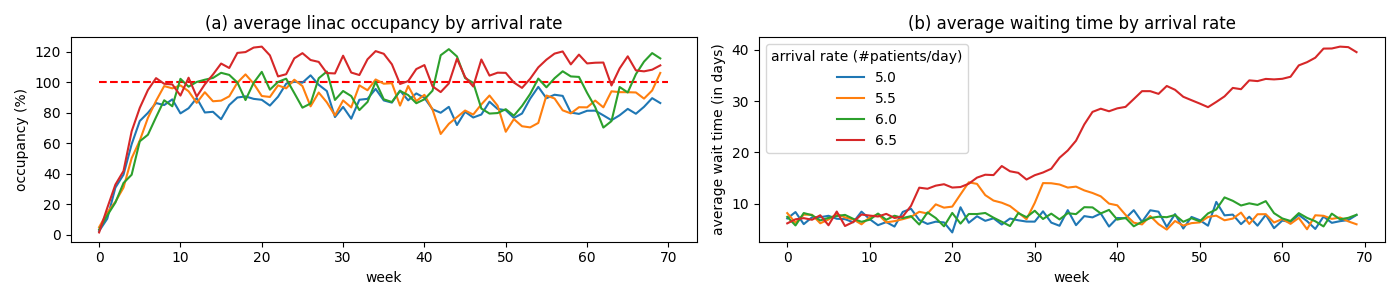}
	\caption{Simulation on 4 linacs with arrival rates of 5.0, 5.5, 6.0 and 6.5. }
	\label{fig:4linacs}
\end{figure}

After identifying a reasonable range of arrival rates, we performed experiments using our realistic instance settings. 
Figure \ref{fig:4linacs_5.0} shows the results for 4 linacs with an arrival rate of 5.0. 
Observe from the figure that in an ideal situation where all future arrivals are known in advance, the overdue time is insignificant in all patient categories. The greedy heuristic and the three batch scheduling strategies are similar in both waiting time and overdue time. Overdue treatments mostly occur in P2 and P3 patients. The prediction-based approach shows a slightly better average waiting time and overdue time in all patient categories. It also results in less deviation in waiting time and overdue time between patients. For each priority group, we performed an one-way ANOVA test. The results confirm that the difference between the tested algorithms are statistically significant, with p-values in all four priority groups less than \textit{1E-5}. The post hoc paired t-tests confirm that the results for the prediction-based approach for P1, P2, and P4 are substantially different from the results for the other four approaches. 
However, in priority group P3, there is no significant difference between the prediction-based approach and the weekly-IP-based approach (p-value $= 0.33$).\looseness=-1

Figure \ref{fig:4linacs_6.0} shows results for the same hospital size in a more crowded scenario with an arrival rate of 6.0. 
We can observe from the figure that the more crowded a hospital is, the better the prediction-based approach performs compared to batch scheduling strategies and the greedy heuristic. 
The most significant difference is for P2 patients, the most challenging category for other algorithms. 
As P2 patients account for almost $30\%$ of patients, and their treatment deadlines are urgent (only 3 days after their admission date), it is difficult to decide how much space should be reserved for their arrivals when scheduling curative patients. 
Despite having less information than batch scheduling strategies, the prediction-based approach demonstrates an ability to learn good decision-making from offline schedules. Specifically, it learns to dynamically delay treatment for curative patients in order to make space for palliative treatments based on the present state of the schedule. 
This explains the lower overdue rates associated with prediction-based scheduling in palliative patients.\looseness=-1

\begin{figure}[h]
	\centering
	\begin{minipage}{.48\textwidth}
		\centering
		\includegraphics[width=\linewidth]{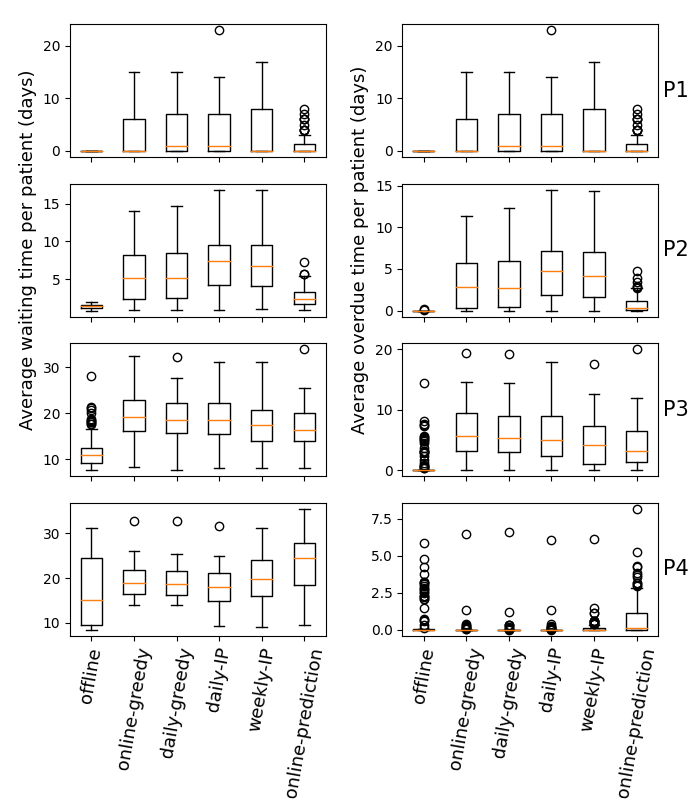}
		\caption{Simulation on 4 linacs - arrival rate 5.0. }
		\label{fig:4linacs_5.0}
	\end{minipage}%
	\hspace{0.1cm}
	\begin{minipage}{.48\textwidth}
		\centering
		\includegraphics[width=\linewidth]{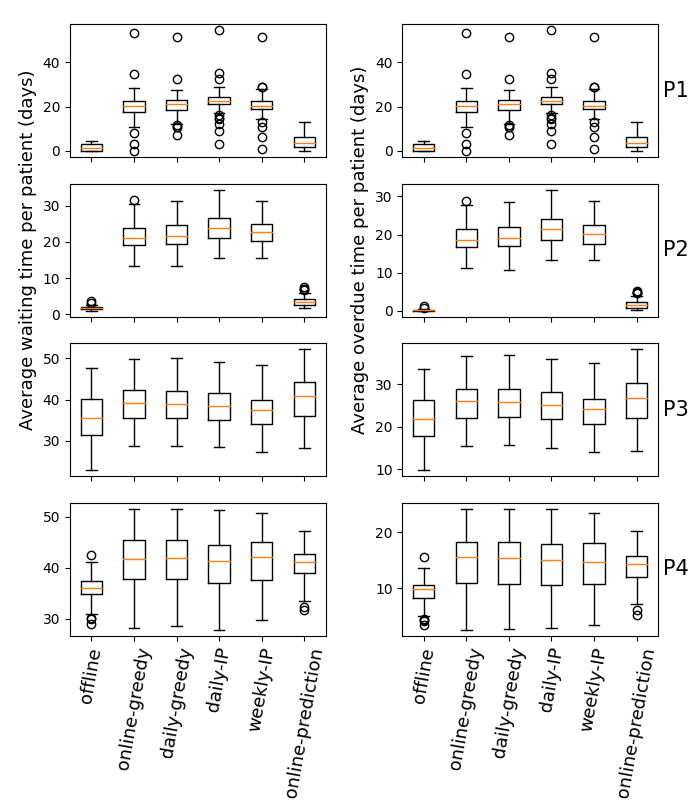}
		\caption{Simulation on 4 linacs - arrival rate 6.0. }
		\label{fig:4linacs_6.0}
	\end{minipage}%
\end{figure}

For the next experiment, we tested the algorithms on a much larger hospital with 8 linacs. 
The reasonable arrival rates suggested by the capacity simulations are less than 12 patients per day, as an arrival rate of 13.0 causes the weekly average waiting time to increase over time (Figure \ref{fig:8linacs}). 
We tested two settings with arrival rates of 10.0, and 12.0. The results are shown in Figures \ref{fig:8linacs_10}, and \ref{fig:8linacs_12}, respectively. We derive two observations from these results. First, the experiment matches our earlier observation from experiments using 4 linacs that the larger and more crowded a hospital is, the better the prediction-based approach performs compared to other scheduling strategies. Second, our approach scales well on large instances. To our knowledge, most experiments in the literature tested their ideas on much smaller hospital sizes. Furthermore, most real hospitals do not have more than 8 linacs operating full-time. In large hospitals, as the number of linacs and patients grows, the scheduling task becomes extremely challenging for most algorithms. Our prediction-based approach offers a fast online approach with superior results.\looseness=-1

\begin{figure}[h!]
\centering
\includegraphics[width=\linewidth]{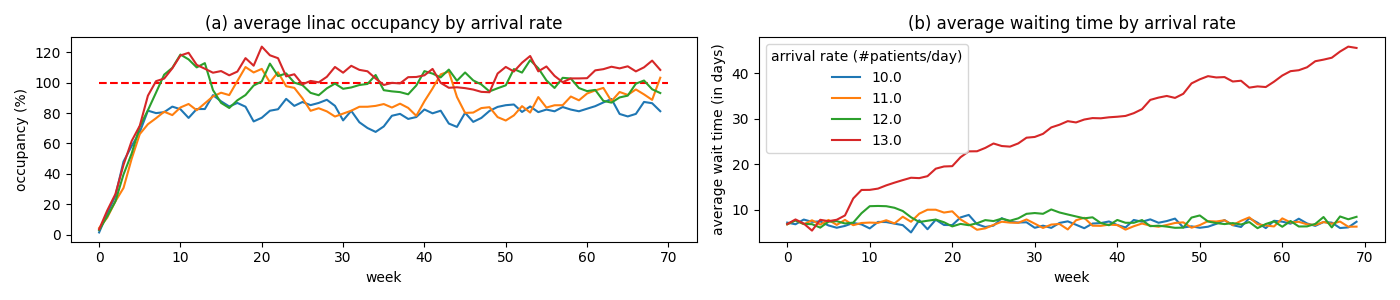}
\caption{Capacity simulation on 8 linacs with arrival rates of 10, 11, 12 and 13. Starting from an arrival rate of 13, the average waiting time of patients increases over time. } 
\label{fig:8linacs}
\end{figure}

\begin{figure}[h!]
\centering
\begin{minipage}{.48\textwidth}
\centering
\includegraphics[width=\linewidth]{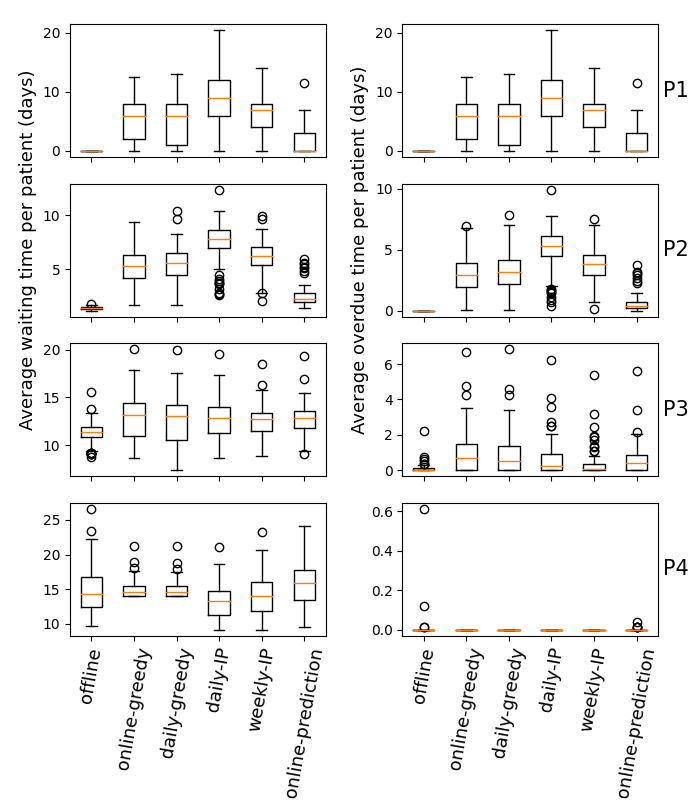}
\caption{Simulation on 8 linacs - arrival rate 10.0. }
\label{fig:8linacs_10}
\end{minipage}%
\hspace{0.1cm}
\begin{minipage}{.48\textwidth}
\centering
\includegraphics[width=\linewidth]{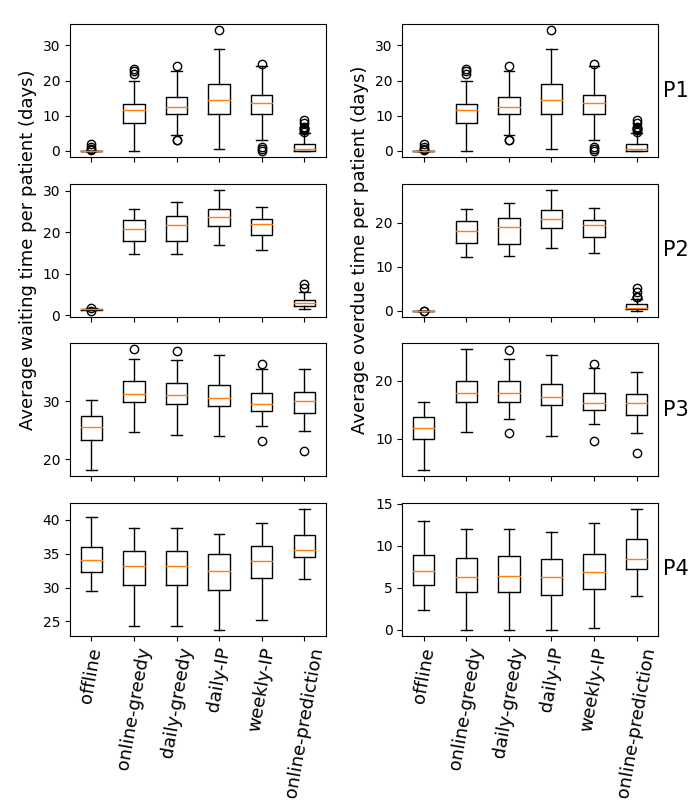}
\caption{Simulation on 8 linacs - arrival rate 12.0. }
\label{fig:8linacs_12}
\end{minipage}%
\end{figure}

\subsection{Numerical results on real instances}
\label{sec:exp_realins}

In a real-world scenario, we might observe a high fluctuation in arrival rate. In this experiment, we aim to test the model's sensitivity to real patient flow with an unstable arrival rate. 
We extract the real patient flow for the period from May 2018 to June 2019.  
As mentioned in Section \ref{sec:data_overview}, CHUM's facility consists of 12 linacs, 3 of which are for special treatments. 
Therefore, we remove all patients bound to special linacs, which accounts for $9\%$ of patients. 
Among the remaining 9 linacs, only 5 generic linacs operate full-time (10 hours a day). 
The remaining 4 linacs operate only half of each working day on average due to a lack of technicians. 
We construct real instances using the extracted patient flow and 7 linacs operating full-time. 
Then we use the first 100 days to initialize a partially-filled schedule using the method described in Section \ref{sec:exp_datagen}. 
Finally, the remaining days of data are used as the patient flow for the instance. The patient flow of the real instance consists of patient arrivals for 180 days, from October 2018 to June 2019. 
In Figure \ref{fig:fit_poisson}, we plot the real arrival rates at CHUM for the period from May 2018 to June 2019. The blue line represents the real daily arrivals of patients, and the red line represents the average arrival rate of a 10-day interval prior to a given date. Observe from the figure that the average arrival rate fluctuates significantly, reaching $7.75$ patients/day at its lowest, and reaching $13.42$ patients/day at its highest.\looseness=-1

\begin{figure*}[]
\centering
\includegraphics[width=0.8\linewidth]{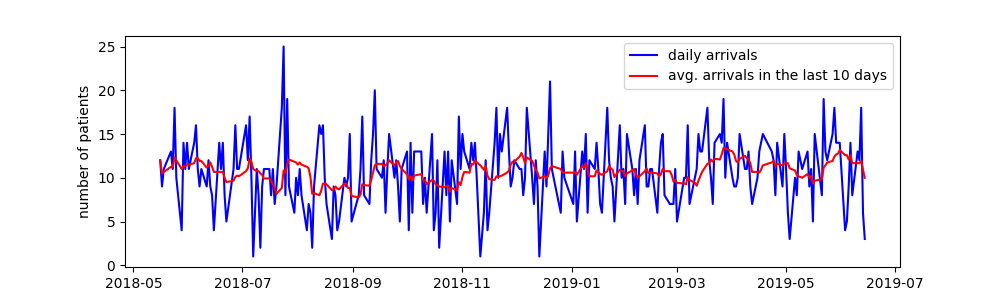}
\caption{Real patient flow at CHUM (from May 2018 to June 2019).}
\label{fig:fit_poisson}
\end{figure*}

The regression model for the real instance is trained on the setting with 7 linacs and an arrival rate of 10.1, which is the average arrival rate in the first 3 months of the data. 
Similar to Section \ref{sec:exp_result_generateddata}, we ran all proposed scheduling strategies on 100 test instances. The results are presented in Figure \ref{fig:realins_testing}. Similar to the results from generated instances in Section \ref{sec:exp_result_generateddata}, the results here show that the prediction-based approach outperforms other approaches in palliative patients while maintaining a comparable waiting time and overdue time for curative patients. 
Observed from Figure \ref{fig:fit_poisson} that the arrival rate fluctuates by time in real life, we also train a regression model on a variable arrival rate. The training instances are generated with a dynamic arrival rate that changes every 10 days within the range of $[\lambda - 1.5, \lambda + 1.5]$, with $\lambda$ being the average arrival rate ($10.1$ patients per day in our case). The parameters of the data generator are chosen based on the observation of the real arrival rate's fluctuation (the red line in Figure \ref{fig:fit_poisson}). The 10 days-interval and the deviation of $\pm1.5$ in arrival rate mimic the fluctuation of the real data while preventing sudden changes in arrival rate. 
	The online prediction-based approach using the variable arrival rate is labeled as \textit{prediction-based*}. As the fluctuated model is trained on a different dataset, we do not include its results in Figure \ref{fig:realins_testing}. 
\looseness=-1

\begin{figure}[h!]
\centering
\begin{minipage}{.45\textwidth}
\centering
\includegraphics[width=\linewidth]{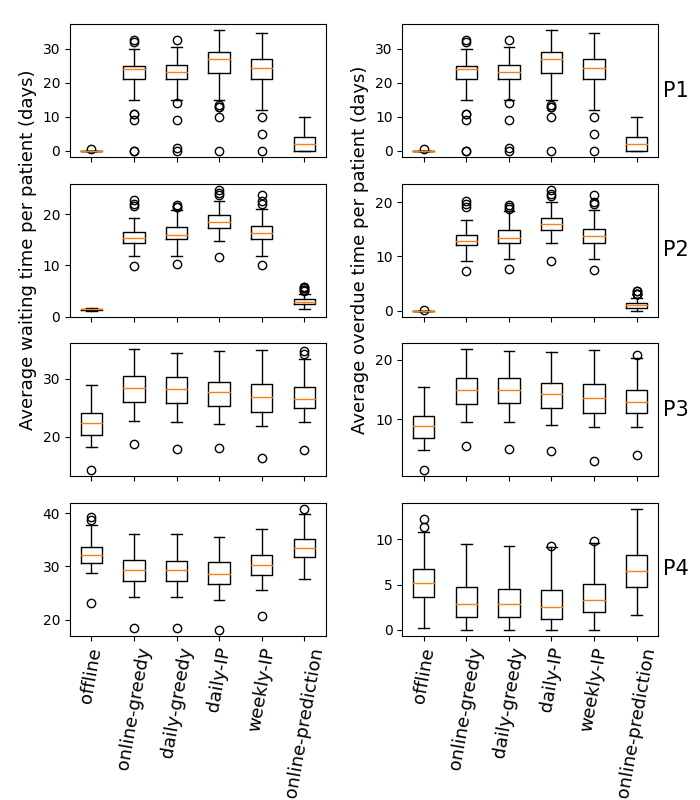}
\caption{Simulation on 7 linacs, arrival rate 10.1. }
\label{fig:realins_testing}
\end{minipage}%
\hspace{0.1cm}
\begin{minipage}{.52\textwidth}
\centering
\includegraphics[width=\linewidth]{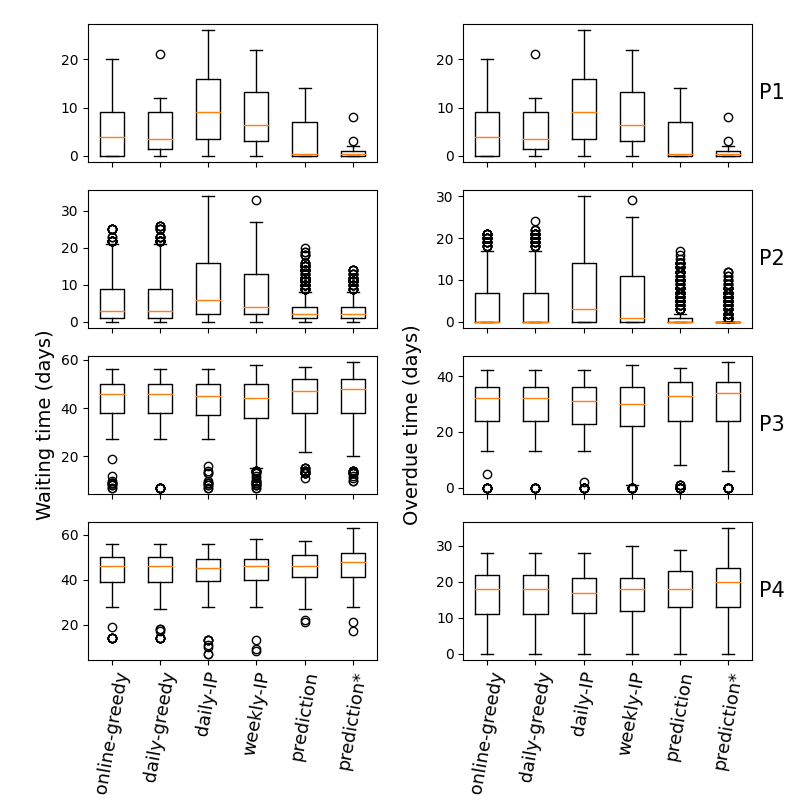}
\caption{Simulation on the real instance.} 
\label{fig:realins_90}
\end{minipage}%
\end{figure}

Then we test the trained regression model on the real patient flow. 
Due to the long simulation period, the result of the offline scheduling is not available. 
	We first analyse the results given by the prediction-based approach with a stable arrival rate. 
The results presented in Table \ref{tab:realins_90} and Figure \ref{fig:realins_90} show that the prediction-based approach results in the lowest average overdue time in both P1 (3.29 days) and P2 (1.99 days). 
The greedy algorithms (both online and daily) perform relatively well on palliative patients, but result in 2 extra days of average overdue time compared to the prediction-based approach. Statistical tests (one-way ANOVA and post-hoc paired t-test) confirm the differences between the results of the tested algorithms on P2 (all p-values are less than $7E\mbox{-}31$) are statistically significant. 
On P1, even though the one-way ANOVA finds no significant differences in the results (p-value$ = 0.073$), the statistical differences between the prediction-based approach and the other four approaches are significant. The paired t-tests between the results on P1 given by the prediction-based approach and other approaches 
all have p-values of less than $0.05$. 
In curative patients, we observe little difference in the results. 
The average overdue time in P3 given by the prediction-based approach is about one day higher than other approaches. This small delay is compensated for by the large reduction in overdue treatments of palliative patients. 

Looking at the results given by the prediction-based approach trained on a variable arrival rate (prediction-based*), we observe that the approach reduces even further the overdue time in palliative patients ($1.21$ days in P1 and $1.0$ in P2, compared to $3.29$ days in P1 and $1.99$ days in P2 given by the model trained on a stable arrival rate). The box-plots in Figure \ref{fig:realins_90} shows that there are also less outliers in overdue time of palliative patients. However, there is a trade-off of a higher overdue time in curative patients ($0.27$ days higher in P3 and $0.86$ days higher in P4) compared to the prediction-based approach with a stable arrival rate.  
We conclude that the prediction approach shows its robustness in real-life scenarios where there is a high fluctuation in patient arrivals. This allows us to avoid the cost of retraining the model, as that would be necessary only when hospitals upgrade their infrastructure or experience a significant change in arrival rates.\looseness=-1

\begin{table}[]
		\centering
		\small
	\begin{tabular}{c|c|ccccc|ccccc}
		\multirow{2}{*}{\begin{tabular}[c]{@{}c@{}}Scheduling\\ strategy\end{tabular}} & \multicolumn{1}{c}{\multirow{2}{*}{\begin{tabular}[c]{@{}c@{}}avg.\\ occupancy\end{tabular}}} & \multicolumn{5}{c}{Waiting time (days)} & \multicolumn{5}{c}{Overdue time (days)} \\
		& \multicolumn{1}{c}{} & \multicolumn{1}{c}{overall} & \multicolumn{1}{c}{P1} & \multicolumn{1}{c}{P2} & \multicolumn{1}{c}{P3} & \multicolumn{1}{c}{P4} & \multicolumn{1}{c}{overall} & \multicolumn{1}{c}{P1} & \multicolumn{1}{c}{P2} & \multicolumn{1}{c}{P3} & \multicolumn{1}{c}{P4} \\ \hline
		online-greedy & 97.45 & 33.02 & 5.14 & 6.13 & 3.67 & 44.02 & 17.80 & 5.14 & 3.91 & 29.74 & 16.18 \\
		daily-greedy & 97.51 & 32.91 & 6.00 & 6.23 & 43.48 & 43.80 & 17.71 & 6.00 & 3.99 & 29.58 & 16.00 \\
		daily-IP & 97.72 & 33.53 & 9.79 & 9.63 & 42.87 & \textbf{43.44} & 18.25 & 9.79 & 7.15 & 28.93 & \textbf{15.65} \\
		weekly-IP & 97.61 & 33.04 & 7.86 & 7.72 & \textbf{42.42} & 44.10 & 17.76 & 7.86 & 5.37 & \textbf{28.51} & 16.19 \\
		prediction-based & 97.14 & 32.93 & \textbf{3.29} & \textbf{4.05} & 44.21 & 44.94 & 17.69 & \textbf{3.29} & \textbf{1.99} & 30.22 & 16.96\\
		prediction-based* & 96.89 &	33.02&\textbf{	1.21}&	\textbf{3.03}&	44.46&	45.80&	17.79&	\textbf{1.21}&	\textbf{1.00}&	30.49&	17.82
	\end{tabular}
\caption{Average waiting time of patients for different scheduling strategies on the real instance.}
\label{tab:realins_90}
\end{table}

\subsection{Explainable decision-making with SHAP value}
In this section, we use SHAP (SHapley Additive exPlanations) to analyse the regression model used in our prediction-based approach, and show how it can assist explainability in scheduling decisions. SHAP values represent the relative impact of a variable on the outcome. 
To demonstrate the methodology for our analysis, we use the regression model for the instance setting with 6 linacs and an arrival rate of 9.0. 
We first look at a global interpretation of the model, which gives us an overview of the impact each feature has on the decision. 
A beeswarm plot (Figure \ref{fig:global_shap}) is an information-dense summary of how top features impact the model's output. Each point corresponds to an  observation in the training set. The y-axis indicates the feature name, and the higher a feature is, the more impact it has on the output. The x-axis indicates the SHAP values of the data points. All the values to the left of the centre line represent the observations that shift the output in the negative direction (reducing the prediction of waiting time), while the points on the right shift the prediction in a positive direction (increasing the waiting time). The colors represent the values of the features from low to high. 
Observe from Figure \ref{fig:global_shap} that for the given model, the number of sections has the highest impact on the decision. A high number of sections has a positive impact on the output, i.e. a high number of sections is likely to result in a large waiting time and vice versa. The contribution of the remaining features can be interpreted similarly, e.g, a large due day has a high and positive impact on the waiting time (patients with high due days have more waiting times), while a low remaining capacity on day 49 results in a higher waiting time. 
\looseness=-1 

\begin{figure*}[]
	\centering
	\includegraphics[width=\linewidth]{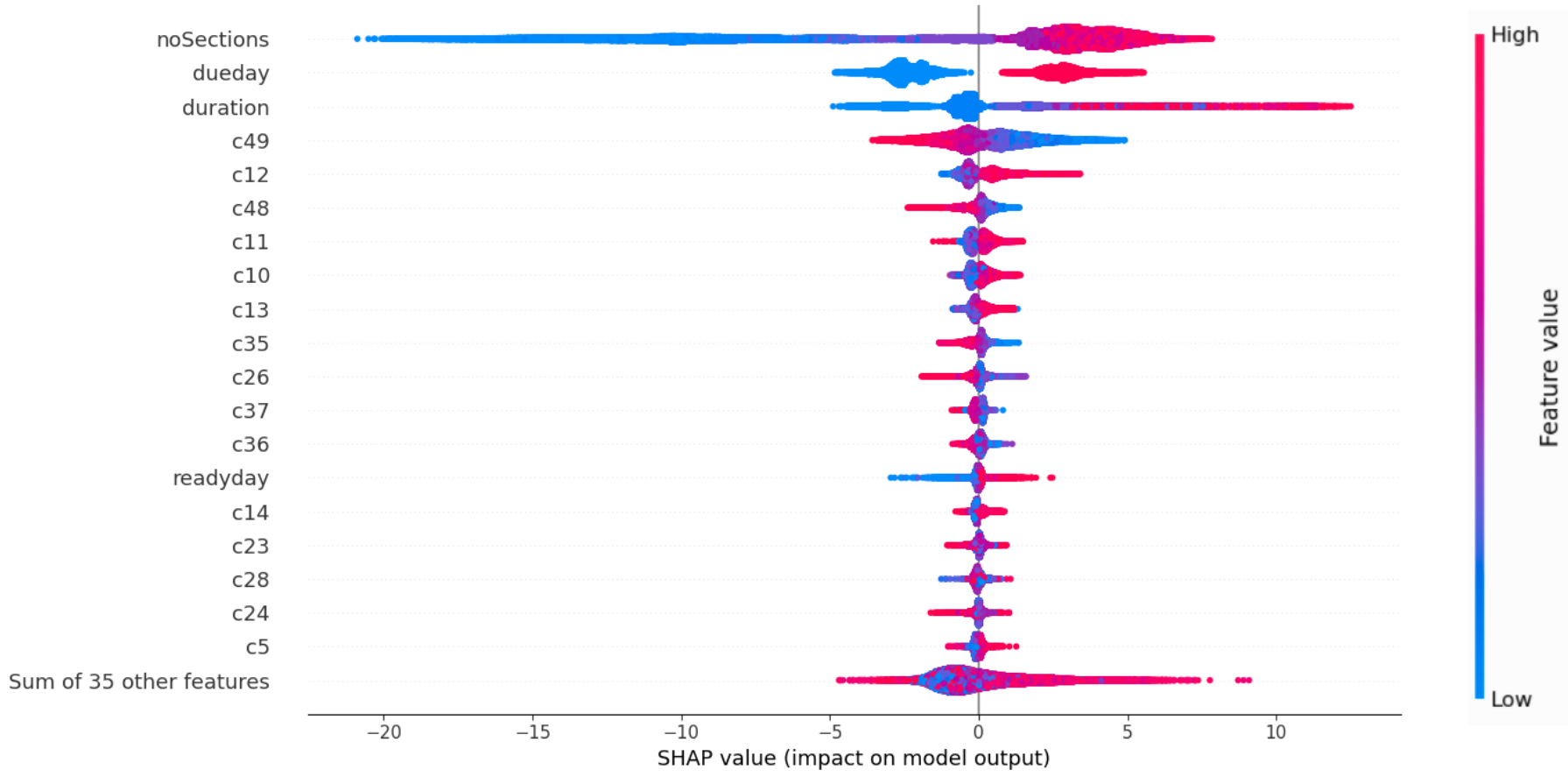}
	\caption{Beeswarm plot for the instance setting with 6 linacs and an arrival rate of 9.0.}
	\label{fig:global_shap}
\end{figure*}

To explain an individual scheduling decision we need to look at a plot that offers local interpretability. 
A waterfall plot visualizes how the feature values contribute to the final prediction of an individual observation. Two examples of waterfall graphs for two different patients can be found in Figures \ref{fig:waterfall_100} and \ref{fig:waterfall_5000}.
The x-axis shows the value of the output (number of waiting days in our problem). $E[f(x)]$ is the base value, which is generally the average output of the training set. To explain a decision, we start from the base value at the bottom of the plot, then add up the contribution of each feature (positive for red and negative for blue) one row at a time, until we reach the actual model output $f(x)$ at the top of the plot. 
The gray text before the feature names shows the value of each feature for this sample. Thus, the waterfall plot explains how much each feature contributes to the final decision.\looseness=-1

\begin{figure}[h!]
	\centering
	\begin{minipage}{.48\textwidth}
		\centering
		\includegraphics[width=\linewidth]{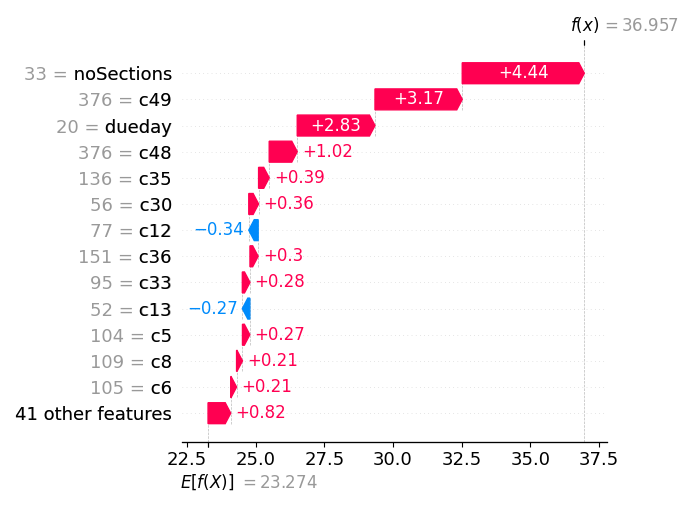}
		\caption{The waterfall plot for a P4-patient.}
		\label{fig:waterfall_100}
	\end{minipage}%
	\hspace{0.1cm}
	\begin{minipage}{.48\textwidth}
		\centering
		\includegraphics[width=\linewidth]{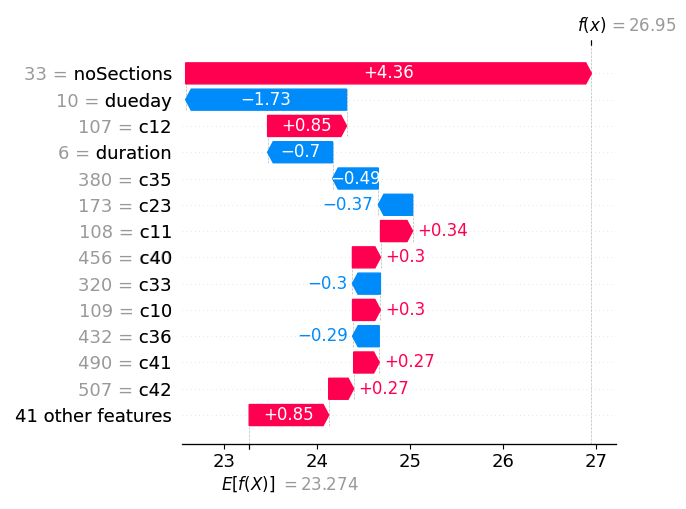}
		\caption{The waterfall plot for a P3-patient.}
		\label{fig:waterfall_5000}
	\end{minipage}%
\end{figure}

\section{Discussion and future work}
\label{sec:discussion}
In this section, we discuss the potential impact and challenges of our implementation in the real world, as well as possible future research directions. 
Currently, our industrial partner, the oncology department of CHUM, is collaborating with Gray\footnote{\url{https://www.graysuite.com/}} (a patient scheduling start-up), to develop an application for their scheduling tasks. The goal of the collaboration is to automate the scheduling tasks that are currently done manually. We have been actively collaborating with Gray from the beginning. The scheduling system is based on a modified IP model. It was deployed in early 2022, and Gray is constantly using feedback from CHUM to improve the system. 
The system is still myopic, i.e. it does not take stochastic information into account when making scheduling decisions. Overdue treatments, especially in P2 and P3 patients, remain a big challenge. 
We hope to be able to conduct tests using the predictive model in the second half of 2022. 
Supporting explainable decisions is another advantage of our approach. 
From our experience deploying the system, building confidence is the most challenging task, as it is difficult to convince the scheduling staff to ``trust'' the algorithm, especially when decisions are human-sensitive. Therefore, transparent scheduling systems are highly appreciated by the industry.\looseness=-1

When implementing a machine learning-based approach for a real world application, it is important to consider computing costs. 
In our approach, the most expensive cost is associated with computing offline solutions. For our largest instance size (8 linacs, arrival rates of 12.0), the average run time is 47,522 seconds ($\sim$13 hours). 
Generating offline solutions for 500 instances requires about 7000 hours of CPU time. It is our opinion that this is both reasonable and feasible for large hospitals, and can be done easily with a computer cluster or server. Additionally, the regression model adapts well to fluctuation in arrival rates and does not need to be retrained often, as shown in Section \ref{sec:exp_realins}. 
However, for hospitals of larger size than the one in this study, generating offline solutions might become a problem. Moving forward, we plan to apply unsupervised learning to train the regression model and overcome this obstacle.\looseness=-1 

Patient scheduling consists of allocation scheduling and appointment scheduling. 
 Solving both problems at once is difficult due to the high complexity. Studies that consider both problems either (1) decompose the problem into 2 phases \citep{pham2021two, vieira2020radiotherapy} where appointment scheduling is solved either in the second phase or as a post-processing step, or (2) use (meta)heuristics \citep{maschler2016particle,maschler2018particle,vogl2019scheduling}. 
 However, those approaches do not guarantee the best performance. It is an open question if compromising some objectives in  allocation scheduling would lead to a better solution in appointment scheduling in terms of the number of fixed appointments that were rescheduled, patient satisfaction, etc. 
 We plan to investigate this in future research.\looseness=-1

\color{black}
\section{Conclusions}
\label{sec:conclusion}
Scheduling radiotherapy treatment is difficult due to the stochastic arrival rate of patients with different priorities who require multiple appointments. In this paper, we proposed an online machine learning-based approach that makes scheduling decisions dynamically based on the present allocation profile of the hospital. 
The algorithm was evaluated on a rolling horizon and compared with several myopic approaches, both heuristic-based and IP-based. 
The test instances were generated based on real data from CHUM to simulate different hospital sizes and crowding levels. 
The results show that our prediction-based approach outperforms other approaches. The prediction-based approach results in a lower average overdue time and lower average waiting time for palliative patients when compared with myopic strategies, while maintaining the same performance for curative patients. 
Our approach also scales decently with instance size, and we successfully solve instances with up to 8 linacs. 
We proved the robustness of our prediction-based approach in a real problem instance from CHUM, despite the high fluctuation of arrival rates observed in reality. Finally, we demonstrated how our approach supports explainability and interpretability.\looseness=-1


\bibliographystyle{abbrvnat}
\bibliography{myrefs}

\appendix

\section{Training regression models}
\label{app:trainingresult}

Before training the regression model, we test the collinearity of the features to verify if there are highly correlated features that should be removed. 
A correlation heatmap can be found in Figure \ref{fig:heatmap}. 
We can see from the figure that there is a certain correlation between the available capacity of consecutive days (the capacity reduces gradually over time). We plotted only one-third of the considered days. However, we consider that information is relevant to the decision.\looseness=-1   

\begin{figure*}[]
	\centering
	\includegraphics[width=1.0\linewidth]{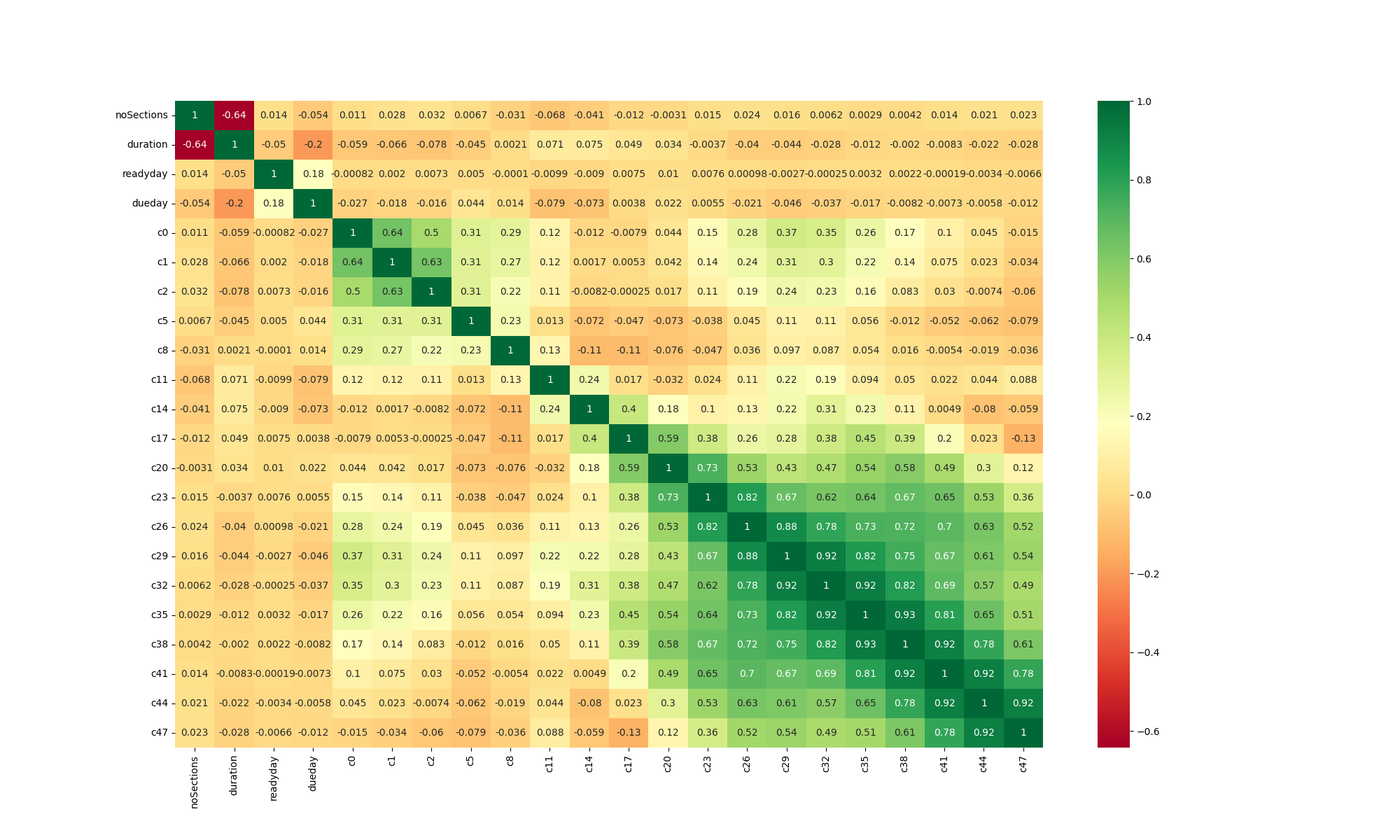}
	\caption{The correlation heatmap of input features.}
	\label{fig:heatmap}
\end{figure*}

\color{black}

After confirming the independence of input features, different regression models are tested. 
We present the results on different instance settings, from easiest to hardest. 
Eight regression models were tested. For each model and instance setting, we report the training time, mean squared error (MSE), mean absolute error (MAE), and R-squared value on both the training set and the testing set.\looseness=-1   

Three instance settings were tested, from the easiest to the most difficult: 

\begin{itemize}
\item 4 linacs, arrival rate of 5.0
\item 6 linacs, arrival rate of 9.0
\item 8 linacs, arrival rate of 12.0
\end{itemize}
The training results for the three settings are presented in Table \ref{tab:trainingresults_4linacs}, \ref{tab:trainingresults_6linacs}, and \ref{tab:trainingresults_8linacs}, respectively. 
Random Forest and XGBoost consistently gave the best results on both the training set and the testing set. However, Random Forest requires more time for training, and it performs worse in difficult settings, i.e., when the size of the training test increases (69.12, 212.88, and 294.18 seconds in instance settings of 4, 6, and 8 linacs, respectively). Meanwhile, XGBoost takes only 6.56 seconds in 4 linacs and 2.86 seconds in 8 linacs. 
Random Forest gives the lowest error rates, but the MAE in the testing set given by Random Forest and XGBoost are very close. A lower differences between the error rates in the training test and the setting test from XGBoost also indicate that it is less prone to overfitting compared to Random Forest. The R-squared values given by XGBoost in the three test cases are approximate $90\%$, which shows that the data fits the regression model well. Based on the above analysis, we choose XGBoost as the regression model for our algorithm.\looseness=-1   

\begin{table}[]
	\centering
	\small
	\begin{tabular}{l|r|rrr|rrr}
		\multirow{2}{*}{} & \multicolumn{1}{l}{} & \multicolumn{3}{c}{Training} & \multicolumn{3}{c}{Testing} \\
		& \multicolumn{1}{c}{Training time} & \multicolumn{1}{l}{MSE} & \multicolumn{1}{c}{MAE} & \multicolumn{1}{c}{R-squared} & \multicolumn{1}{c}{MSE} & \multicolumn{1}{c}{MAE} & \multicolumn{1}{c}{R-squared} \\ \hline
		MLP & 29.7 & 6.28 & 1.8 & 0.81 & 5.94 & 1.76 & 0.82 \\
		SGD & 0.15 & 12.82 & 2.78 & 0.61 & 12.38 & 2.75 & 0.62 \\
		Lasso & 0.96 & 12.79 & 2.78 & 0.61 & 12.36 & 2.75 & 0.62 \\
		ElasticNet & 0.72 & 13.06 & 2.81 & 0.6 & 12.62 & 2.77 & 0.61 \\
		Decision Tree & 1.22 & 3.6 & 0.59 & 0.89 & 9.85 & 1.8 & 0.7 \\
		Random forest & 69.12 & \textbf{0.6} & \textbf{0.49} & 0.98 & 4.43 & 1.35 & 0.86 \\
		XGBoost & 6.56 & 1.5 & 0.85 & 0.95 & 4.3 & \textbf{1.34} & 0.87
	\end{tabular}
\caption{Training results for instance setting with 4 linacs, and an arrival rate of 5.0.}
\label{tab:trainingresults_4linacs}
\end{table}

\begin{table}[]
	\centering
	\small
	\begin{tabular}{l|r|rrr|rrr}
		& \multicolumn{1}{l}{} & \multicolumn{3}{c}{Training} & \multicolumn{3}{c}{Testing} \\
		& \multicolumn{1}{c}{Training time} & \multicolumn{1}{l}{MSE} & \multicolumn{1}{c}{MAE} & \multicolumn{1}{c}{R-squared} & \multicolumn{1}{c}{MSE} & \multicolumn{1}{c}{MAE} & \multicolumn{1}{c}{R-squared} \\	\hline
		MLP & 139.93 & 10.55 & 2.34 & 0.85 & 10.31 & 2.29 & 0.85 \\
		SGD & 0.65 & 24.6 & 3.68 & 0.66 & 23.71 & 3.61 & 0.65 \\
		Lasso & 1.65 & 24.31 & 3.66 & 0.66 & 23.4 & 3.59 & 0.66 \\
		ElasticNet & 1.13 & 24.51 & 3.66 & 0.66 & 23.56 & 3.58 & 0.66 \\
		Decision Tree & 3.54 & 8.11 & 1.16 & 0.89 & 14.42 & 2.13 & 0.79 \\
		Random forest & 212.88 & \textbf{0.92} & \textbf{0.56} & 0.99 & \textbf{6.7} & \textbf{1.55} & 0.9 \\
		XGBoost & 2.37 & 3.45 & 1.26 & 0.95 & 6.78 & 1.64 & 0.9
	\end{tabular}
\caption{Training results for instance setting with 6 linacs, and an arrival rate of 9.0.}
\label{tab:trainingresults_6linacs}
\end{table}

\begin{table}[]
	\centering
	\small
	\begin{tabular}{l|r|rrr|rrr}
		& \multicolumn{1}{l}{} & \multicolumn{3}{c}{Training} & \multicolumn{3}{c}{Testing} \\
		& \multicolumn{1}{c}{Training time} & \multicolumn{1}{l}{MSE} & \multicolumn{1}{c}{MAE} & \multicolumn{1}{c}{R-squared} & \multicolumn{1}{c}{MSE} & \multicolumn{1}{c}{MAE} & \multicolumn{1}{c}{R-squared} \\	\hline
		MLP & 159.35 & 7.27 & 1.91 & 0.87 & 7.46 & 1.94 & 0.87 \\
		SGD & 0.89 & 18.93 & 3.25 & 0.67 & 19.34 & 3.27 & 0.67 \\
		Lasso & 1.3 & 18.88 & 3.23 & 0.67 & 19.26 & 3.25 & 0.67 \\
		ElasticNet & 0.57 & 19.03 & 3.23 & 0.67 & 19.45 & 3.26 & 0.67 \\
		Decision Tree & 4.82 & 5.49 & 0.9 & 0.9 & 10.11 & 1.76 & 0.83 \\
		Random forest & 294.18 & \textbf{0.57} & \textbf{0.44} & 0.99 & \textbf{4.36} & \textbf{1.27} & 0.93 \\
		XGBoost & 2.86 & 2.57 & 1.08 & 0.95 & 4.54 & 1.37 & 0.92
	\end{tabular}
\caption{Training results for instance setting with 8 linacs, and an arrival rate of 12.0.}
\label{tab:trainingresults_8linacs}
\end{table}

\section{Additional results on generated instances}
\label{app:6linacs}
In this section, we provide the results for additional instance settings with a hospital size of 6 linacs.
The realistic arrival rates identified through the capacity simulations are between 7.0 and 9.0 (see Figure \ref{fig:6linacs}). 

\begin{figure}[h]
	\centering
	\includegraphics[width=\linewidth]{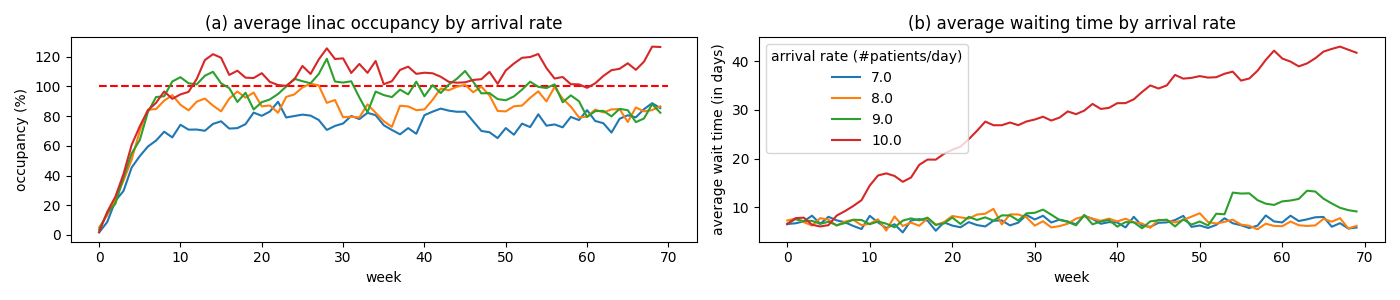}
	\caption{Capacity simulation on 6 linacs with arrival rates of 7.0, 8.0, 9.0 and 10.0. Starting from an arrival rate of 10.0, the average waiting time of patients increases over time. }
	\label{fig:6linacs}
\end{figure}

The waiting times and overdue times given by the proposed algorithms using those settings are shown in Figures \ref{fig:6linacs_7.0}, and \ref{fig:6linacs_9.0}, respectively. 
Observe that the differences between the algorithms are more profound than in the previous smaller settings with 4 linacs, especially when the arrival rate increases. 
The prediction-based approach offers better waiting times and overdue times for palliative patients (both P1 and P2) than scheduling methods, while maintaining a comparable waiting time for curative patients. 
Similar to the case with 4 linacs, the more crowded a hospital is, the better the prediction-based approach performs over other approaches.
Another observation from the boxplots is that the average waiting time (and overdue time) of P4 patients in greedy heuristic and batch scheduling is lower than the others, which also leads to higher overdue time in palliative patients. 
This shows that the greedy heuristic and batch scheduling are not effective in delaying curative treatments to accommodate palliative ones. The results are in line with the results in Section \ref{sec:exp_result_generateddata}, and confirm that the prediction-based approach performs better in large and crowded hospitals.\looseness=-1   

\begin{figure}[h]
	\centering
	\begin{minipage}{.48\textwidth}
		\centering
		\includegraphics[width=\linewidth]{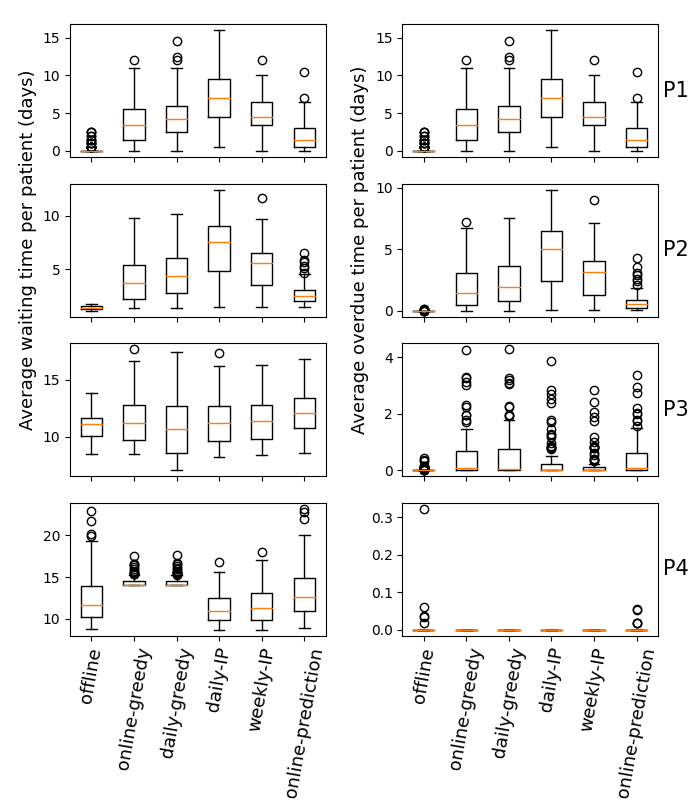}
		\caption{Simulation on 6 linacs - arrival rate 7.0. }
		\label{fig:6linacs_7.0}
	\end{minipage}%
	\hspace{0.1cm}
	\begin{minipage}{.48\textwidth}
		\centering
		\includegraphics[width=\linewidth]{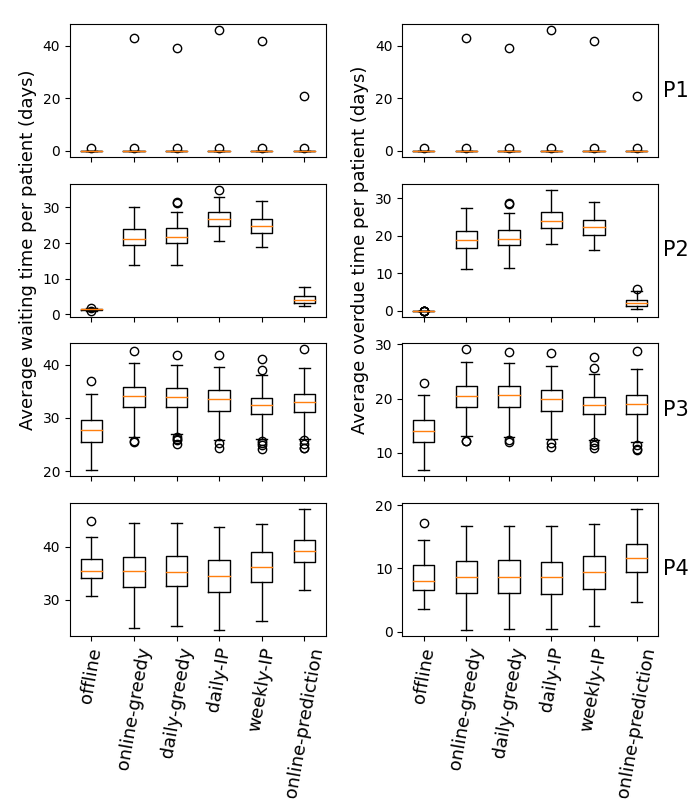}
		\caption{Simulation on 6 linacs - arrival rate 9.0. }
		\label{fig:6linacs_9.0}
	\end{minipage}%
\end{figure}



\end{document}